\documentclass[letterpaper]{article} 
\usepackage{aaai2026}  
\nocopyright
\usepackage{times}  
\usepackage{helvet}  
\usepackage{courier}  
\usepackage[hyphens]{url}  
\usepackage{graphicx} 
\usepackage{natbib}  
\usepackage{caption} 
\usepackage{algorithm}
\usepackage{algorithmic}

\usepackage{amsmath}
\usepackage{amsfonts}
\usepackage{booktabs} 
\usepackage{multirow} 
\usepackage{siunitx}  
\usepackage{tabularx} 
\usepackage{ragged2e} 
\usepackage{subcaption}
\usepackage{mathtools}
\usepackage{newfloat}
\usepackage{listings}
\DeclareCaptionStyle{ruled}{labelfont=normalfont,labelsep=colon,strut=off} 
\floatstyle{ruled}
\newfloat{listing}{tb}{lst}{}
\floatname{listing}{Listing}
\title{REINA: Regularized Entropy Information-Based Loss for Efficient \\ Simultaneous Speech Translation}
\author{
Nameer Hirschkind\equalcontrib, Joseph Liu\equalcontrib, Xiao Yu, Mahesh Kumar Nandwana
}
\affiliations{
    Roblox\

    \{nhirschkind, josephliu, xyu, mnandwana\}@roblox.com
}

\usepackage{bibentry}

\begin{document}

\maketitle

\begin{abstract}
Simultaneous Speech Translation (SimulST) systems stream in audio while simultaneously emitting translated text or speech. Such systems face the significant challenge of balancing translation quality and latency. We introduce a strategy to optimize this tradeoff: wait for more input only if you gain information by doing so. Based on this strategy, we present Regularized Entropy INformation Adaptation (REINA), a novel loss to train an adaptive policy using an existing non-streaming translation model. We derive REINA from information theory principles and show that REINA helps push the reported Pareto frontier of the latency/quality tradeoff over prior works.
Utilizing REINA, we train a SimulST model on French, Spanish and German, both from and into English. Training on only open source or synthetically generated data, we achieve state-of-the-art (SOTA) streaming results for models of comparable size. We also introduce a metric for streaming efficiency, quantitatively showing REINA improves the latency/quality trade-off by as much as 21 percent compared to prior approaches, normalized against non-streaming baseline BLEU scores.
\end{abstract}


\section{Introduction}


Simultaneous Speech Translation (SimulST) involves real-time translation of speech in one language into text in another. 
This extends the simpler speech-to-text-translation (S2TT) task, which involves translation with the full context of an entire speech clip. While S2TT allows for offline applications, conversational environments such as voice or video chat necessitate SimulST models to facilitate real-time communication across language barriers.

Recently, End-to-end (E2E) S2TT models have largely superseded traditional cascaded approaches, which link separate Automatic Speech Recognition (ASR) and Machine Translation (MT) systems. 
E2E models mitigate error propagation and reduce latency by directly mapping source speech to target text~\cite{communication2023seamlessmultilingualexpressivestreaming,peng2024owsm,whisper_paper,hirschkind24_interspeech}. 

Rather than training SimulST models from scratch, most works take advantage of advances in S2TT research by adapting a non-streaming S2TT model into a SimulST model~\cite{communication2023seamlessmultilingualexpressivestreaming,dig_sst_2024,streamspeech,papi2024streamatt}. To transition from non-streaming S2TT to SimulST introduces the challenge of balancing translation quality and latency. This requires a policy to decide whether to wait for more input (READ) or generate output (WRITE)~\cite{gu-etal-2017-learning}. This problem is particularly difficult in the translation setting because different languages can have different word orderings, requiring differing amounts of context before a suitable translation can occur. Many approaches have been formulated to determine this READ/WRITE policy, from baking the policy into the model architecture itself via monotonic attention mechanisms ~\cite{arivazhagan-etal-2019-monotonic-milk, Ma2020Monotonic, communication2023seamlessmultilingualexpressivestreaming}, or having a separate module to dictate the policy \cite{dig_sst_2024}. However, these approaches suffer from issues including poor translation quality compared to non-streaming models and expensive, numerically unstable training~\cite{communication2023seamlessmultilingualexpressivestreaming, dig_sst_2024}.

In this paper, we address the problem of efficiently training high quality SimulST models. The major contributions of this paper can be summarized as follows:

\textbf{New Policy Training Technique.} We propose Regularized Entropy INformation Adaptation (REINA), a new technique for policy training that can efficiently convert non-streaming Speech-to-Text Translation (S2TT) models into simultaneous S2TT (SimulST) models. 
REINA is guided by an approximation of mutual information derived from the S2TT model’s log probabilities on partial versus full audio and is shown to produce higher quality policies than existing methods.

\textbf{Trained with open-source data.} We train an E2E S2TT model with REINA on 130k hours of open-source data. Based on empirical studies, this model achieves SOTA streaming translation performance.

\textbf{Streaming efficiency evaluation metric.} We propose a new evaluation metric to better compare SimulST models. This metric normalizes the streaming translation quality against the performance of the underlying non-streaming model, allowing for a fairer assessment of the capabilities of the streaming policy itself.

\section{Related Work}



While learning adaptive policies for SimulST is a fairly new research area, it builds on a rich body of non-streaming S2TT work. In this section, we outline these S2TT foundations and then discuss SimulST policies.

\textbf{Training Speech to Text Translation Models} 
The literature around S2TT contains many large-scale, powerful models including Whisper \cite{whisper_paper}, SeamlessM4T \cite{communication2023seamlessmultilingualexpressivestreaming}, Canary \cite{puvvada2024less}, and the Open Whisper-style Speech Model (OWSM) \cite{peng2024owsm}. 
These models vary in architecture (e.g., Whisper's Transformer, OWSMv3.1's E-Branchformer) and training data scale, ranging from Canary's ~86k hours (leveraging pseudo-labels) to Whisper's ~680k hours of web data and SeamlessM4T's ~600k hours of synthetically aligned data.

Due to the relative scarcity of parallel ST data compared to ASR or MT corpora, multi-task learning (MTL) is widely adopted~\cite{ye2022cross, dig_sst_2024, communication2023seamlessmultilingualexpressivestreaming}. Auxiliary tasks like ASR and neural machine translation (NMT) are jointly trained with S2TT to improve representations and leverage abundant text or speech data. Furthermore, contrastive learning techniques, such as in ConST \cite{ye2022cross}, encourage similarity between corresponding speech segments and their transcriptions. There is a notable gap in the literature between industry work leveraging massive proprietary datasets and less resourced research making heavy use of MTL to get the most out of smaller data scales. As OWSM \cite{peng2024owsm} bridges this gap for the non-streaming setting, we aim to do the same for SimulST. We are one of the first SimulST works to leverage large-scale open source data that we train on with an MTL framework including MT and ASR tasks.


\textbf{Streaming Policy Learning} 
Transitioning S2TT models to SimulST introduces the challenge of learning a READ/WRITE policy that balances translation quality and latency.
Fixed policies like wait-k are simple to implement but are usually suboptimal due to the mismatch between the sampling rate of the input audio frames and the frequency of outputted words~\cite{ma-etal-2020-simulmt}. That being said, recent works like SimulS2S-LLM~\cite{deng2025simuls2s} have used wait-k with some success. Adaptive polices based on heuristics such as attention matrix weights as seen in EdAtt~\cite{papi-etal-2023-attention} also been proposed, which are usually better.

On the other hand, adaptive learnable policies dynamically adjust decisions based on context.
Prior works have integrated the policy within the model architecture, such as Transducer models~\cite{graves2012sequencetransductionrecurrentneural, xue2022large}, which inherently support streaming via monotonic alignment, or models using monotonic attention mechanisms like MMA~\cite{arivazhagan-etal-2019-monotonic-milk} or EMMA~\cite{communication2023seamlessmultilingualexpressivestreaming}.
Monotonic alignment methods afford greater expressivity, but they tend to be excessively expensive to compute at train time and suffer from both poor numerical stability and difficulty in converging. Our preliminary investigations validate this claim.

Other SimulST works avoid complex, explicit policies, instead generating aligned data with which to directly train SimulST models \cite{labiausse2025high-hibiki, fu2025efficient, deng2025simuls2sllmunlockingsimultaneousinference}. These works often use existing models such as NMT models \cite{labiausse2025high-hibiki} or LLM's~\cite{fu2025efficient, deng2025simuls2sllmunlockingsimultaneousinference} as teachers to create synthetically aligned data for streaming training. Such models can afford simpler architectures without policy networks.
That said, SimulST models deriving their policies from generated data are often limited in their streaming performance based on the quality of the teacher model.

\textbf{Explicit policy training}
Other adaptive strategies decouple the policy from the translation model~\cite{dig_sst_2024, gu-etal-2017-learning, streamspeech}. Some leverage signals from pre-trained offline models, such as using reinforcement learning (RL) to directly optimize the quality-latency trade-off~\cite{gu-etal-2017-learning}. 
Although these methods simplify the learning problem by decoupling the policy from the translation model, they require explicit supervision from a suitable metric and are often suboptimal. 
RL is hard to stabilize and efficiently train, especially in cases like SimulST, with no guarantee of convergence~\cite{gu-etal-2017-learning}.

Closely related to our work is the divergence-guided approach of DiG-SST~\cite{dig_sst_2024}. DiG-SST trains a lightweight policy module using the expected divergence between output distributions conditioned on partial versus complete input, estimated from a non-streaming S2TT model. This approach is efficient to train and directly optimizes for the SimulST task. Nevertheless, DiG-SST's formulation fails to make use of valuable information from ground truth labels when computing divergence scores. In REINA, we propose an improved formulation of a similar concept, yielding better streaming results.


\section{Model}
\begin{figure*}[t]
\centering
  \includegraphics[width=0.85\linewidth]{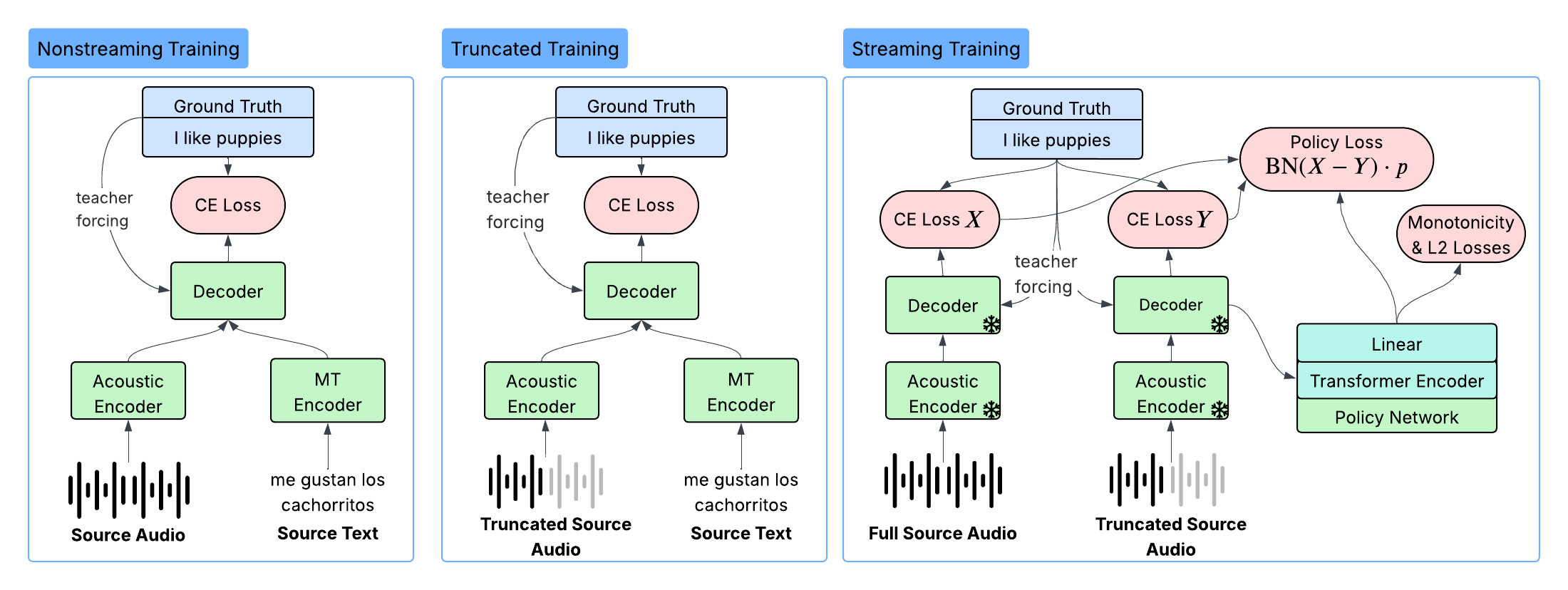}
  \caption {Non-streaming and streaming training procedures for REINAStream. For non-streaming training we use a trainable MT encoder to train on parallel NMT data. During streaming training we a) pass a full audio and truncated audio through the model, b) compute the cross-entropy (CE) loss of each, c) predict a policy using the policy network on top of the partial-audio output of the decoder, and finally d) calculate the REINA loss using the CE terms and policy predictions.}
  \label{fig:wholearch}
\end{figure*}

In this section, we introduce REINAStream, a low latency SimulST model trained on large-scale, open source data. We illustrate our architecture and loss functions in figure \ref{fig:wholearch}.

\subsection{Policy Learning}
To learn an effective READ/WRITE policy, we introduce a new loss function: Regularized Entropy INformation Adaptation (REINA). REINA enables us to adapt a non-streaming speech translation model into a streaming, SimulST model with minimal extra training.

First, we outline the problem more formally. 
Suppose that we are translating an input audio stream $a$ into a target language with a streaming chunk size of $j$ frames. Given a partial audio recording $a_t$ at frame $t$ and previously emitted tokens $s_1, s_2, \ldots, s_n$, we need to decide whether to produce token $s_{n+1}$ (WRITE) or wait for another audio chunk (READ). If we READ, we consume another frame of audio, giving us $a_{t+1}$, whereas if we WRITE, we gain a token, yielding the same audio $a_t$ but tokens $s_1, \ldots, s_n, s_{n+1}$. Our policy must learn to make READ/WRITE decisions that maximize translation quality while minimizing the latency with which we emit each token. This recursive setup appears to lend itself to a dynamic programming type of optimization, as in Seamless \cite{communication2023seamlessmultilingualexpressivestreaming} or the Transducer architecture \cite{graves2012sequencetransductionrecurrentneural}. However, in practice, optimizing over all possible READ/WRITE sequences results in expensive, numerically unstable training.

Instead, we start from a core idea: we should wait for more audio (i.e. READ) if and only if we gain information by doing so. We formalize this notion using mutual information theory. 
Given audio $a$ of length $T$ and ground truth translation token sequence $S=(s_1, \ldots, s_N)$, after writing $n < N$ tokens and listening to $t < T$ timesteps of audio, we can express the information gained about the next token $s_{n+1}$ by waiting for the rest of the input audio as 
\begin{align}\mathcal{F}(a, S, n, t) \vcentcolon= I(s_{n+1}; a_T, S_n) - I(s_{n+1}; a_t, S_n) 
\end{align}

 where $I$ is the symbol for mutual information. We can then construct an ideal READ/WRITE policy $\pi_{\alpha}$ on top of this quantity: $\pi_{\alpha}(a, S, n, t)$ returns READ when $\mathcal{F}(a, S, n, t) > \alpha$ and WRITE otherwise. We can then adjust $\alpha$ to control the latency quality tradeoff. This policy READs exactly when the information gained exceeds a given threshold. We can rewrite $\mathcal{F}(a, S, n, t)$ using mutual information equations~\cite{barber2004algorithm} as follows

\begin{align} 
\mathcal{F}&(a, S, n, t) \vcentcolon= I(s_{n+1}; a_T, S_n) - I(s_{n+1}; a_t, S_n) \nonumber \\
&= H(s_{n+1}) - H(s_{n+1} | a_T, S_n) \nonumber \\
&- [H(s_{n+1}) - H(s_{n+1} | a_t, S_n)] \nonumber \\
&= H(s_{n+1} | a_t, S_n) - H(s_{n+1} | a_T, S_n) \nonumber \\
&= \mathbb{E}\left[ \log p (s_{n+1} | a_T, S_n) - \log p (s_{n+1} | a_t, S_n) \right]
\end{align}

Although we might not have access to $\log p (s_{n+1} | a_T, S_n)$ or $\log p (s_{n+1} | a_t, S_n)$, we are able to estimate these via the log-probabilities of the base S2TT model as $\log \hat{p} (s_{n+1} | a_T, S_n)$ and $\log \hat{p} (s_{n+1} | a_t, S_n)$, which are the log-probabilities of the next label token computed when passing the full audio and partial audio through the model respectively. Note that these are also the negatives of the cross-entropy losses obtained when running the model on the full and partial audio.

We now have an estimate $\mathcal{\hat{F}}(a, S, n, t)$. 
However, this estimate cannot be computed during inference time as it requires access to the full target text, so we instead formulate a heuristic to estimate the information gain. 
We train a model parameterized by $\theta$ to estimate a heuristic $q_{\theta}=q(a, S, n, t | \theta)$ that yields policy $\hat{\pi}_{\alpha}$, which returns READ if and only if $q_{\theta}>\alpha$. We train $q_{\theta}$ to strongly correlate with $\mathcal{F}(a, S, n, t)$ by maximizing the covariance between $q_{\theta}$  and the estimate $\mathcal{\hat{F}}(a, S, n, t)$. This gives us the following optimization problem:
\begin{align}
\max_{\theta} \Bigl(&\text{Cov}(q_{\theta} , \mathcal{\hat{F}}(a, S, n, t)) \Bigr) \nonumber  \\
= \max_{\theta} \Bigl( &\mathbb{E}\left[q_{\theta} \cdot \mathcal{\hat{F}}(a, S, n, t)\right] - \nonumber \\
&\mathbb{E}\left[q_{\theta}\right]\cdot \mathbb{E}\left[\mathcal{\hat{F}}(a, S, n, t)\right] \Bigr) 
\end{align}
We can simplify this expression by normalizing the information gain estimate $\mathcal{\hat{F}}(a, S, n, t)$ to have zero mean, making $\mathbb{E}\left[q_{\theta}\right]\cdot \mathbb{E}\left[\mathcal{\hat{F}}(a, S, n, t)\right]$  evaluate to 0. We achieve this at train time by normalizing each batch to have 0 mean and 1 standard deviation, resulting in the final optimization problem:
\begin{align}
\max_{\theta} \Bigl[&\mathbb{E}\left[q_{\theta} \cdot \text{BN}\Bigl(\mathcal{\hat{F}}(a, S, n, t)\Bigr)\right] \Bigr]\nonumber  \\
= \min_{\theta} \Bigl[&q_{\theta} \cdot  \text{BN}  \bigl[  \log \hat{p_t}^{s_{n+1}}  - \log \hat{p_T}^{s_{n+1}} \bigr] \Bigr]
\end{align}
where BN represents our batch normalization. We apply a shorthand $\log \hat{p_T}^{s_{n+1}}= \log \hat{p}(s_{n+1} | a_T, S_n)$ and $\log \hat{p_t}^{s_{n+1}}= \log \hat{p}(s_{n+1} | a_t, S_n)$ for brevity. 

The loss that optimizes the above optimization problem is
\begin{align}
    \mathcal{L}_p = \frac{1}{N}\sum_{n=0}^{N-1} q_{\theta}^n \cdot \text{BN}  \bigl[ \log \hat{p_t}^{s_{n+1}}  - \log \hat{p_T}^{s_{n+1}}) \bigr]
\end{align}

where $q_{\theta}^n = q_{\theta}(a, S, n, t | \theta)$. This loss maximizes the covariance between our estimate of information gain $\hat{\mathcal{F}}$ and our heuristic estimator network $q_{\theta}$. This is the "Policy Loss" in figure \ref{fig:wholearch}. Since $\log \hat{p_t}(s_{n+1})$ and $\log \hat{p_T}(s_{n+1})$ are the negative cross-entropy losses from the decoder given partial and full audio respectively, in the diagram we write $\mathcal{L}_p$ as a difference of cross-entropy loss terms.

To ensure stable training and a reasonable learned policy, we add monotonicity and $L_2$ regularization terms.
First, we note that at inference time, after predicting a READ, we predict no further tokens. Therefore, any WRITEs after a READ are semantically meaningless. 

To better align train and inference, we add a weak monotonicity constraint on $q_{\theta}$ to encourage the probability of READ to increase uniformly across each token sequence $S$:
\begin{align}
\mathcal{L}_m &= \frac{1}{N}\sum_{n=1}^{N} \Bigl[  \text{max} \Bigl( 
\max_{m<n} \{ q_{\theta}^m\} - q_{\theta}^n- \epsilon, 0 \Bigr ) \Bigr]
\end{align}

The loss function $\mathcal{L}_m$ serves as a regularization term to shape the learned heuristic $q_{\theta}$ used in the policy $\hat{\pi}_{\alpha}(a, S, n, t)$. It encourages the sequence of $q_{\theta}$ values for target text tokens, ordered by index $n$, to be approximately non-decreasing ($q_{\theta}^n \ge \max_{m<n} \{ q_{\theta}^m\} - \epsilon$), an inductive bias distinct from merely estimating information gain. This imposed monotonicity biases the policy towards a ``commitment'' behavior, encouraging it to commit to stop predicting new tokens and READ new audio instead once the threshold is crossed for an earlier token.

Finally, we add a simple $L_2$ regularization penalty: $\mathcal{L}_r = \frac{1}{N} \sum_{n=1}^N \left(q_{\theta}^n \right)^2$. We find this is required to prevent $q_{\theta}$ values from exploding to infinity during training. Putting all the terms together, we get the full REINA loss: $\mathcal{L}_{\text{REINA}} = \mathcal{L}_p + \mathcal{L}_m + \lambda \mathcal{L}_r$. In our work, we set $\lambda = 0.05$, but find that final model performance is not very sensitive to changes in $\lambda$. We will cover how we train with $\mathcal{L}_{\text{REINA}}$ in section~\ref{sec:training}.

\subsection{Non-Streaming Architecture}
Next, we describe the architecture of our base non-streaming S2TT model, which is also outlined in figure \ref{fig:wholearch}. The architecture comprises of an acoustic encoder and a text decoder. At train-time, we also use an extra text encoder to facilitate a MT training task.

For the acoustic encoder, we adopt Whisper Medium~\cite{whisper_paper} and do not freeze its weights during training. The randomly initialized transformer decoder performs cross-attention over the acoustic encoder's final layer hidden states and predicts text tokens. We adopt the generic multilingual tokenizer from Mistral 7B~\cite{mistral} but learn our own embedding dictionary. We augment the vocabulary with language ids such as \textless en\textgreater or\textless fr\textgreater  so we can direct the decoder to predict tokens of a specific target language by prefixing the token sequence with the language id. We apply a learned positional encoding similar to Time2Vec~\cite{time2vec} on the acoustic encoder outputs to give the decoder a notion of sequence ordering. The decoder is trained with a cross-entropy loss.

At train-time, to support MT loss calculation, we add a randomly initialized, trainable T5 text encoder~\cite{t5}. We pass source-language text through and then have the decoder cross-attend to the T5 last layer hidden states while predicting target language text. This is facilitates a machine translation task designed to improve the quality of the decoder by making use of paired MT data.

Whisper Medium contains 307M parameters, the text decoder has 101M, and the MT encoder has 38M for a total of 445M trainable parameters at train time and 408M at inference-time. While this is a larger parameter count than academic works like Dig-SST~\cite{dig_sst_2024}, StreamSpeech~\cite{streamspeech}, or Stream-Att~\cite{papi2024streamatt}, it is still much smaller than most industry systems like SeamlessM4T~\cite{communication2023seamlessmultilingualexpressivestreaming} or Hibiki~\cite{labiausse2025high-hibiki}. 
While higher parameter counts demonstrably increase translation quality, they also make it difficult to train and deploy models to large numbers of users in the wild. We target the middle ground between small and large systems in literature, yielding a model with both the translation quality and computational efficiency to be usable in real-world chat settings.
 
\subsection{Streaming Architecture}
We augment the non-streaming model with a policy network that makes binary READ/WRITE decisions for each output token. This is the architecture component that REINA serves to train. The policy network is a small (6M parameter) transformer encoder applied to the last layer hidden states from the decoder. We apply a single linear layer on top of the transformer with output dimension 1 and sigmoid activation in order to make the binary READ/WRITE decisions. We apply a causal attention mask on the policy network at both train and inference time.

\subsection{Training} \label{sec:training}

We train REINAStream in 3 stages: 1) Learn non-streaming S2TT 2) Adapt to truncated audios 3) Learn a streaming policy.

In the first stage, we train on several tasks at once in order to effectively leverage available data to train the speech translation model. All data samples contain some subset of the following information: source language $l^s_i$, source language audio $a_i$, source language transcription $T^s_i$, target language $l^t_i$, and target language transcription $T^t_i$. Using this data, we have three training tasks.

\textbf{ASR} For samples with $(a_i, l^s_i, T^s_i)$, we pass $a_i$ through the acoustic encoder, then decode to tokens in source language $l^s_i$. We compute cross-entropy loss using label transcript $T^s_i$ yielding loss $\mathcal{L}_{asr}$.

\textbf{NMT} For samples with $(T^s_i, l^t_i, T^t_i)$, we pass the source transcript through the T5 text encoder and then decode into target language $l^t_i$. We compute the cross-entropy loss using label transcript $T^t_i$ yielding loss $\mathcal{L}_{nmt}$.

\textbf{S2TT} For samples with $(a_i, l^t_i, T^t_i)$, we pass $a_i$ through the acoustic encoder, then decode into target language $l^t_i$. We compute cross-entropy loss using label transcript $T^t_i$ yielding loss $\mathcal{L}_{s2tt}$.



For the first stage training, we mix data supporting all tasks into every batch and minimize a sum of all the losses: $ \mathcal{L} = \mathcal{L}_{asr} + \mathcal{L}_{nmt} + \mathcal{L}_{s2tt}$.

Before training the policy network on top of the base model, we add a second step to ensure a higher quality estimation of $\log \hat{p}(s_{n+1} | S_n, a_t)$ with partial audios $a_t$ for the policy loss stage. In this phase, we fine-tune the speech translation model on randomly truncated audios using the same loss function $\mathcal{L}$.

Lastly, we train the policy network by minimizing $\mathcal{L}_{\text{REINA}}$ and freezing all other parameters. We only train on S2TT data samples as our goal is to learn a policy best for streaming speech translation rather than streaming ASR or MT.

\begin{table*}[htbp] 
\centering
\fontsize{7}{8}\selectfont
\setlength{\tabcolsep}{5pt} 
\begin{tabular}{@{}ll*{10}{r}@{}}
\toprule
\multirow{3}{*}{Split} & \multirow{3}{*}{Dataset} &
\multicolumn{4}{c}{Source Language: en} &
\multicolumn{2}{c}{Source Language: de} &
\multicolumn{2}{c}{Source Language: es} &
\multicolumn{2}{c}{Source Language: fr} \\
\cmidrule(lr){3-6} \cmidrule(lr){7-8} \cmidrule(lr){9-10} \cmidrule(lr){11-12}

& & \multicolumn{1}{c}{Target: de} & \multicolumn{1}{c}{Target: es} & \multicolumn{1}{c}{Target: fr} & \multicolumn{1}{c}{Target: en} & 
\multicolumn{1}{c}{Target: en} & \multicolumn{1}{c}{Target: de} & 
\multicolumn{1}{c}{Target: en} & \multicolumn{1}{c}{Target: es} & 
\multicolumn{1}{c}{Target: en} & \multicolumn{1}{c}{Target: fr} \\ 

\midrule


\multirow{4}{*}{Train} & MLS
& 13,789 & 13,785 & 13,787 & 41360
& 1,637 & 1,637
& 713 & 713
& 984 & 984  \\

& Must-C
& 386 & 476 & 468 & 1330
& {-} & {-} 
& {-} & {-} 
& {-} & {-} \\


& CVSS-C
&{-}  &{-}  &{-}  & {-} 
& 184 & 184 
& 113 & 113 
& 264 & 264\\


& MOSEL
&{-} & {-} & {-} & 19,245
&{-} & 22,804
&{-} & 19,373
&{-} & 22,835 \\
\cmidrule(lr){1-12}

\multirow{2}{*}{Dev} & Must-C
& 2.47 & 2.49 & 2.49 & {-}
& {-} & {-}
& {-} & {-}
& {-} & {-} \\

& CVSS-C
& {-} & {-} & {-} & {-}
& 21 & {-}
& 22 & {-}
& 22 & {-} \\
\cmidrule(lr){1-12}

\multirow{2}{*}{Test} & Must-C
& 4 & 4 & 4 &{-}
&{-} &{-}
& {-}&{-}
&{-} &{-} \\


& CVSS-C
& {-} & {-} & {-} & {-}
& 22 & {-}
& 23 & {-}
& 23 & {-} \\
\bottomrule
\end{tabular}

\caption{Consolidated Dataset Hours (Source Audio) by Split and Dataset, Grouped by Source and Target Language.}
\label{tab:dataset_hrs}

\end{table*}

\subsection{Data}
We seek to bridge the gap between models trained on large-scale, proprietary datasets and those trained on small-scale open-source data. We leverage a variety of publicly available data sources plus synthetic data generation to produce a large-scale training set. 

In this iteration, we focus only on en$\xrightarrow{}${de, fr, es} and {de, fr, es}$\xrightarrow{}$en language directions because of their data availability. In future work, we plan to expand to lower resourced languages. For audio datasets, we draw from Multilingual Librispeech (MLS) \cite{pratap2020mls}, Mosel \cite{gaido-etal-2024-mosel}, CVSS-C \cite{jia2022cvss}, MUST-C \cite{di-gangi-etal-2019-mustc}. We list details on these datasets in table \ref{tab:dataset_hrs}. We further augment the MLS dataset by translating its transcripts using an in-house NMT model to produce S2TT data to train on. We also augment our dataset with text-to-text MT training data from CCMatrix \cite{schwenk-etal-2021-ccmatrix}. We use 10M samples per language pair from CCMatrix for a total of 60M samples.

\subsection{Inference Policy}
We use streaming beam search to perform inference. We split input audios into $0.25$s chunks and inference the model on all audio up to the current chunk in sequence.

We first pick a policy threshold $\alpha$ to control the quality-latency tradeoff while streaming. Each iteration of the search, we run the policy network on all beams. When a policy prediction is less than $\alpha$ we perform a READ, adding the tokens predicted in that beam to the running list of hypotheses and setting the the beam's logprob to 0 so it isn't used to continue the search. If the total number of beams that have encountered a READ action exceeds the beam size times a patience factor, or all beams hit READ at once, we end the search. After ending the search, we return the hypothesis with the highest average log probability.

Once we have reached the end of the input audio, we stop using the policy network and beam search until we reach the EOS token, using the same patience factor logic to decide when to end the search.

\section{Experiments}
In this section, we detail how we ran our experiments and present our results compared with existing work.

\subsection{Experimental Setup}

To train REINAStream, we follow the procedure outlined in Section~\ref{sec:training}, beginning with the non-streaming S2TT model. The model's text decoder is a 16-layer transformer with a model dimension of 512, 8 attention heads, a feedforward multiplier of 4, and uses label smoothing and dropout rates of 0.1. We train this initial model (Stage 1) for 5 days on 24 A100-80G GPUs using an AdamW optimizer~\cite{adamw}. The training configuration includes a fixed learning rate of $10^{-4}$, a weight decay of $10^{-4}$, and gradient clipping set to 10.0, with an effective batch size of 768. The data for this stage is a mixture of MUST-C (ratio 1), CVSS (ratio 1), MLS (en $\rightarrow$ X) (ratio 2), and MLS (X $\rightarrow$ en), CCMatrix, and Mosel (all at ratio 4). For the second truncated stage (Stage 2), we use the exact same architecture and training configuration for 2 days, but train on a data mix of 20\% full audios and 80\% randomly truncated audios.

Lastly, we perform the REINA policy training (Stage 3). The REINA policy network is a 2-layer transformer with a 512 dimension embedding, 4 attention heads, and a feedforward multiplier of 4. It is regularized with a monotonicity loss with $\epsilon= 0.5$ and an L2 Regularization weight $\lambda=.05$. For this stage, we use an inverse square root learning rate scheduler with 5k warmup steps. We also adjust the dataset mixing ratios, increasing the ratios for MUST-C and CVSS to 2, and both directions of MLS to 6. This final training stage completes 20 epochs in under 12 hours.

We also re-implement DiG-SST's divergence-based loss from \cite{dig_sst_2024} based on the description in the paper and train the policy network with that loss using the same configuration as we train REINA.

We evaluate on language pairs $\{$fr, de, es$\}\xrightarrow{}$ en on CVSS-C, and on language pairs en $\xrightarrow{}\{$fr, de, es$\}$ on MUST-C. We use a beam size of 3 with no length penalty, a streaming chunk size of 0.25s, and a patience factor of 3. We sweep across several thresholds for the policy network in order to measure our model's tradeoff between latency and accuracy. Choosing the right thresholds to obtain an informative sweep is done purely through trial and error.

We compare to existing works on each dataset that we believe to be at or near to state-of-the-art at the time of writing. On MUST-C, we compare to Dig-SST \cite{dig_sst_2024}, the work in literature closest to ours, and another strong SimulST competitor called DiSeg \cite{diseg}. On CVSS-C, we compare to StreamSpeech \cite{streamspeech} and SimulS2S-LLM~\cite{deng2025simuls2s}, a recent work showing strong simultaneous S2ST performance on CVSS with a small model. As StreamSpeech only reports ASR-BLEU whereas we report text BLEU, we are unable to make a fair comparison. That said, StreamSpeech is a stronger system than most in the literature on CVSS-C, and we outperform SimulS2S-LLM which outperforms StreamSpeech, so we show the results nontheless. 
For all comparisons, we use self-reported results from the original papers.

We also perform several ablations on the REINA training stage of REINAStream to demonstrate the utility of the different parts of the REINA loss and make fairer comparisons. We train five model variants:


\textbf{REINA} Our standard training procedure including all S2TT training datasets with the standard REINA loss function.

\textbf{REINA w/o monotonicity} Just like \textbf{REINA} but without the monotonicity term in the loss. Trained on all S2TT datasets.

\textbf{REINA (MUST-C only)} We train the policy network with the full REINA loss on only the MUST-C dataset for a fairer comparison to Dig-SST.

\textbf{REINA w/o truncated training} Just like \textbf{REINA} but skipping the truncated training stage. Trained on only the MUST-C dataset for comparison to Dig-SST.

\textbf{Dig-SST (Our impl. MUST-C only)} We train our own implementation of the DiG-SST on top of the non-streaming REINAStream model on only the MUST-C dataset.   

\subsection{Evaluation Metrics}
We measure translation accuracy using BLEU as implemented in the SacreBLEU package~\cite{sacrebleu}, Average Lag (AL) and Length-Adaptive Average Lag (LAAL)~\cite{papi-etal-2022-generation}, which we implement ourselves based on the original paper \cite{ma-etal-2020-simuleval}. Most existing works plot AL vs BLEU curves by interpolating between several (AL, BLEU) points generated by evaluating their model with different streaming settings \cite{dig_sst_2024, papi2024streamatt}.
We contend that this evaluation does not sufficiently disentangle a model's non-streaming translation quality from its streaming ability. 
We observe that many comparisons in the literature pitch models as being better at streaming than others due to having a higher BLEU vs AL curve, when in reality the difference may be accounted for entirely by one model having a superior non-streaming BLEU.

We wish to show that REINA adapts S2TT models of any non-streaming performance to SimulST models with minimal degradation in translation quality, and compare against other methods that modify a non-streaming model to be streaming.
To this end, we introduce Normalized Streaming Efficiency (NoSE), a metric to measure streaming performance across the entire quality/latency Pareto frontier normalized by non-streaming translation quality. NoSE is the measure of the area under the AL/BLEU curve (the curve is piecewise-linear so we can compute area exactly), bounded on the left and right by $x<y$ respectively, divided by the area under the non-streaming BLEU line. It is important to note that while we require the $x$ and $y$ bounds to ensure the metric is well-defined, NoSE is heavily dependent on them. For our analysis, we pick the smallest $x$ and largest $y$ for which our work and the works we compare to all have reported values, yielding the widest possible range for which all models have a defined AL/BLEU curve. We recommend that future works using NoSE report the bounds used in their calculations. 

\begin{figure*}[t]
\centering
  \includegraphics[width=1\linewidth]{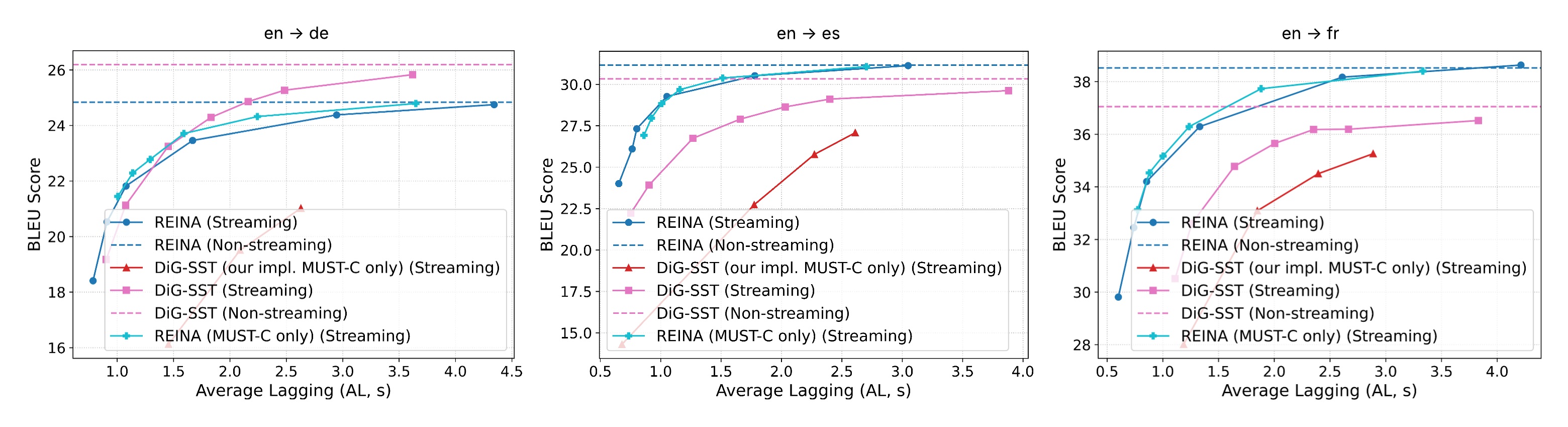}
  \caption {Average Lagging (AL) vs. BLEU score on MUST-C. Horizontal lines represent non-streaming performance.}
  \label{fig:mustc_allplots}
\end{figure*}

\subsection{Results}
We present our NoSe scores in table \ref{tab:combined_results} and selected operating points in table \ref{tab:final_wide_no_op_point} for MUST-C and CVSS results, respectively, with full AL-BLEU tradeoff curves in figure \ref{fig:mustc_allplots}.

 Unfortunately, CVSS-C is not commonly evaluated against in SimulST literature, so we are unable to compare to many other works. However, with an average utterance length of 4.9 seconds, CVSS is the only dataset out of the three comprising primarily shorter audios, which are quite common in conversational SimulST use-cases. This makes it an important benchmark for SimulST systems.

\begin{table}[h] 
\fontsize{6}{7}\selectfont 
\setlength{\tabcolsep}{1.25pt} 
\centering
\begin{tabular}{@{}lccc@{}} 
\toprule
\multicolumn{4}{c}{\textbf{MUST-C}} \\ 
\midrule
Model & en$\rightarrow$de & en$\rightarrow$fr & en$\rightarrow$es \\
\midrule
\itshape Bounds $[x,y]$ & \scriptsize [1.102, 1.965] & \scriptsize [1.187, 1.656] & \scriptsize [1.144, 1.416] \\
\cmidrule(lr){1-4} 
Dig-SST (Original) & 0.888 & 0.903 & 0.879 \\
DiSeg & 0.838 & - & 0.774 \\ 
Dig-SST (Our impl. MUST-C only) & 0.665 & 0.774 & 0.607 \\
EDAtt & 0.704 & - & 0.740\\
REINA (MUST-C only) & \textbf{0.940} & \textbf{0.953} & \textbf{0.960} \\
REINA & 0.925 & 0.944 & 0.952 \\
REINA w/o monotonicity & 0.899 & 0.920 & 0.909 \\
\midrule 
\multicolumn{4}{c}{\textbf{CVSS}} \\ 
\midrule
Model & de$\rightarrow$en & fr$\rightarrow$en & es$\rightarrow$en \\
\midrule
\itshape Bounds $[x,y]$ & \scriptsize [1.955, 5.039] & \scriptsize [1.637, 5.169] & \scriptsize [1.806, 5.587] \\
\cmidrule(lr){1-4} 
StreamSpeech$^*$ & 0.842 & 0.886 & 0.837 \\
REINA w/o monotonicity & \textbf{0.976} & 0.980 & \textbf{0.982} \\
REINA & 0.974 & \textbf{0.983} & 0.981 \\
\bottomrule
\end{tabular}
\caption{NoSE $(\uparrow)$ values for the MUST-C and CVSS-C datasets. Bounds $[x,y]$ for NoSE are specified per language pair. Best values per language pair are bolded. Results for other works are taken from their corresponding papers. $^*$Note StreamSpeech only reports ASR-BLEU, so values for StreamSpeech represent ASR-BLEU scores rather than BLEU. SimulS2S-LLM is not included here as they do not publish their offline BLEU performance.}
\label{tab:combined_results}
\end{table}

\begin{table}[b!]

\centering
\small 
\setlength{\tabcolsep}{1.5pt} 
\fontsize{6.5}{7.5}\selectfont
\begin{tabular}{l rrr rrr rrr}
\toprule
           & \multicolumn{3}{c}{German (En$\rightarrow$De)} & \multicolumn{3}{c}{Spanish (En$\rightarrow$Es)} & \multicolumn{3}{c}{French (En$\rightarrow$Fr)} \\
\cmidrule(lr){2-4} \cmidrule(lr){5-7} \cmidrule(l){8-10}
Model      & AL $\downarrow$ & LAAL $\downarrow$ & BLEU $\uparrow$ & AL $\downarrow$ & LAAL $\downarrow$ & BLEU $\uparrow$ & AL $\downarrow$ & LAAL $\downarrow$ & BLEU $\uparrow$ \\
\midrule
REINA  & 1.01 & 1.10 & 21.44 & 0.86 & 0.94 & 26.92 & 0.77 & 0.87 & 33.13 \\
DiG-SST & 1.08 & --   & 21.13 & 0.90 & --   & 23.92 & 1.11 & --   & 30.51 \\
EDAtt  & 1.04 & 1.20 & 19.10 & 0.95 & 1.24 & 23.00 & --   & --   & --    \\
\midrule
REINA  & 1.59 & 1.67 & 23.71 & 1.16 & 1.24 & 29.68 & 1.24 & 1.32 & 36.29 \\
DiG-SST & 1.45 & --   & 23.25 & 1.27 & --   & 26.74 & 1.26 & --   & 32.61 \\
EDAtt  & 1.34 & 1.46 & 21.60 & 1.28 & 1.52 & 26.60 & --   & --   & --    \\
\midrule
REINA  & 2.24 & 2.30 & 24.32 & 1.51 & 1.57 & 30.38 & 1.88 & 1.93 & 37.73 \\
DiG-SST & 1.83 & --   & 24.29 & 1.66 & --   & 27.90 & 2.00 & --   & 35.65 \\
EDAtt  & 2.26 & 2.33 & 25.60 & 1.52 & 1.74 & 27.80 & --   & --   & --    \\
\midrule
REINA (m)  & 3.65 & 3.67 & 24.79 & 2.70 & 2.73 & 31.07 & 3.33 & 3.35 & 38.40 \\
DIGSST (r) & 3.62 & --   & 25.83 & 2.40 & --   & 29.11 & 3.83 & --   & 36.52 \\
EDAtt (m)  & 2.74 & 2.80 & 26.30 & 2.14 & 2.34 & 29.20 & --   & --   & --    \\
\bottomrule
\end{tabular}
\caption{Comparison of streaming translation models for MUST-C on En$\rightarrow$\{De, Es, Fr\} on various operating points. DiG-SST does not report LAAL values.}
\label{tab:final_wide_no_op_point}
\end{table}

\subsubsection{Quantitative Results}
On both MUST-C and CVSS-C, REINA outperforms all competing methods on all language splits at low latencies. Significantly, this holds for the "MUST-C only" model with policy network trained only on MUST-C, as well as the REINA model trained on all datasets, demonstrating our streaming performance gains are not merely attributable to increased data scale when training the policy, but come from the improved objective. The only exception to this, as seen in table \ref{tab:final_wide_no_op_point}, is German, where DigSST is slightly better at higher latencies than REINA. Still, REINA excels at lower latency streaming, even when its non-streaming BLEU is lower than competitors.

 The MUST-C-only REINA model yields NoSE scores 3.0\% higher than Dig-SST and 8.9\% higher than DiSeg. Our implementation of DiG-SST performs far below every other model in evals, suggesting we missed details during reproduction of results.


\subsubsection{Ablations on Monotonicity Loss}

\begin{figure}[h]
    \centering
    \includegraphics[width=0.75\linewidth]{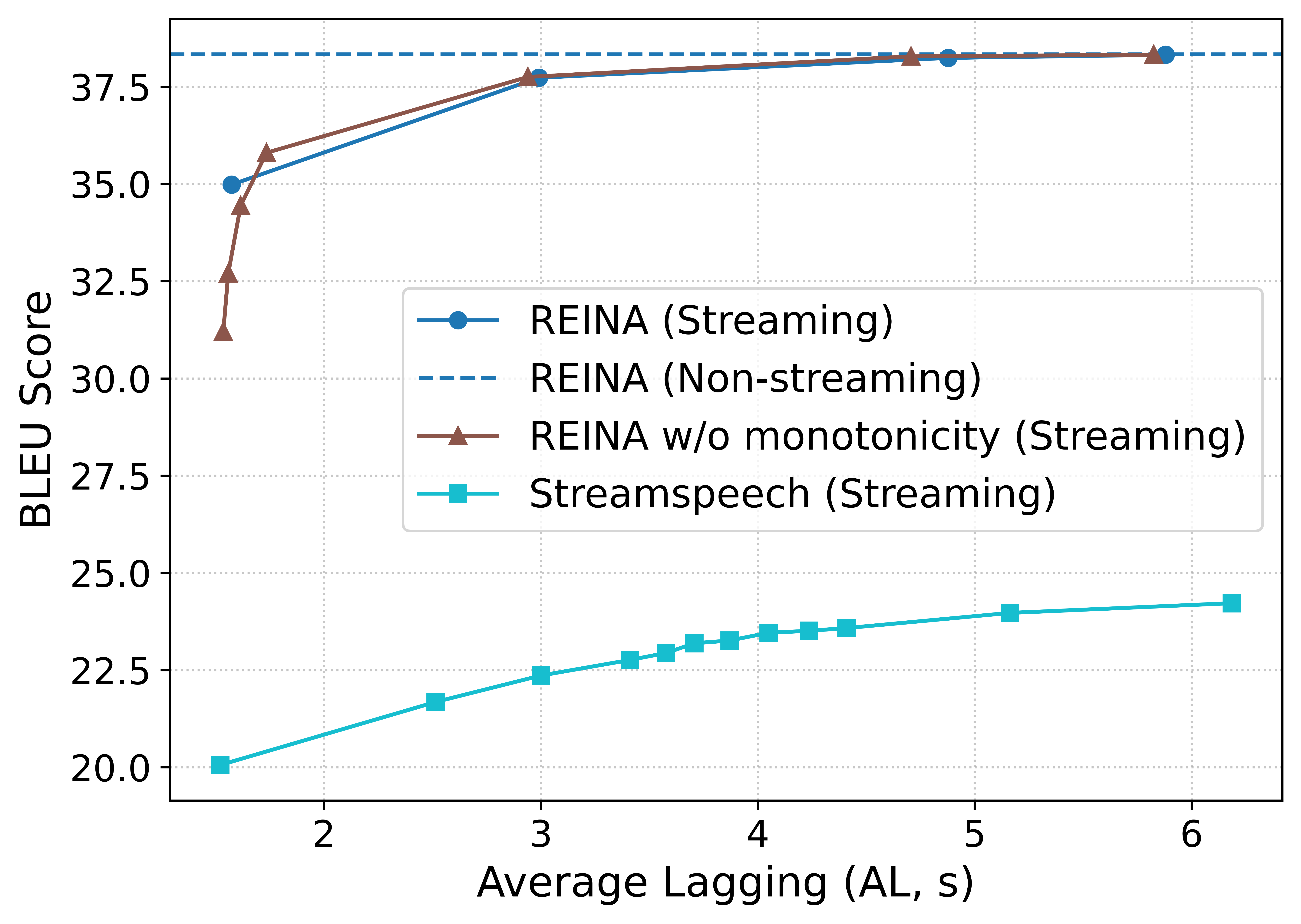}
    \caption{AL/BLEU curve on Es$\rightarrow$En split of the CVSS-C dataset. We report ASR-BLEU only for StreamSpeech.}
    \label{fig:ablate}
\end{figure}


In figure \ref{fig:ablate}, we observe that the REINA model outperforms the REINA w/o monotonicity model exclusively at low latencies, indicating that monotonicity is useful on the most aggressive streaming settings. For example, at about 35 BLEU, AL decreases from 1.95 to 1.57, a 19\% improvement. We hypothesize this is because monotonicity forces the policy to decide on a clear boundary of when to READ when the information gain waffles between timesteps.

\subsubsection{Ablations on Truncated Training}

Table \ref{tab:trunc_ablate} shows the importance of stage 2 training, suggesting that mutual information formulation of REINA requires a good estimate of $\log \hat{p} (s_{n+1} | a_t, S_n)$, as skipping that stage leads to a performance degradation in the trained policy.

\begin{table}[t] 
\fontsize{7}{7.5}\selectfont 
\setlength{\tabcolsep}{1.25pt} 
\centering
\begin{tabular}{@{}lccc@{}} 
\toprule
\multicolumn{4}{c}{\textbf{MUST-C}} \\ 
\midrule
Model & en$\rightarrow$de & en$\rightarrow$fr & en$\rightarrow$es \\
\midrule
\itshape Bounds $[x,y]$ & \scriptsize [1.219, 1.606] & \scriptsize [1.068, 1.468] & \scriptsize [1.276, 1.686] \\
\cmidrule(lr){1-4} 
REINA & 0.932 & 0.942 & 0.971 \\
REINA (No Stage 2 Trunc.) & 0.840 & 0.839 & 0.895 \\
\bottomrule
\end{tabular}
\caption{NoSE $(\uparrow)$ values ablating non-truncated training on MUST-C .}
\label{tab:trunc_ablate}

\end{table}


\section{Conclusion and Future Work}
In this paper, we present a new method for SimulST. We introduce the REINA loss function that enables efficient conversion of non-streaming speech translation models into streaming ones. We conduct extensive experiments over several datasets, showing REINA outperforms the state of the art in SimulST conversion. We also propose a new metric, NoSE, to improve the state of evaluation of SimulST systems. Ultimately, we train a large-scale system entirely on open source or synthetically generated data, encouraging further research into scaling SimulST.

The next step to enable real-time crosslingual interaction is to extend REINAStream into a simultaneous speech to speech translation (SimulS2ST) model. This is achievable by using a high quality, low latency, streaming text-to-speech model as a synthesizer. We are presently working on extending REINAStream to the SimulS2ST use-case.

\bibliography{aaai2026}

\twocolumn[
\begin{center}
    \huge\bfseries Appendix
    \vspace{1.5em}
\end{center}
]
\appendix
\counterwithin{figure}{section}
\counterwithin{table}{section}
\section{NoSE Score}
\label{sec:nose_diag}
\begin{figure}[h]
    \centering
    \includegraphics[width=0.6\linewidth]{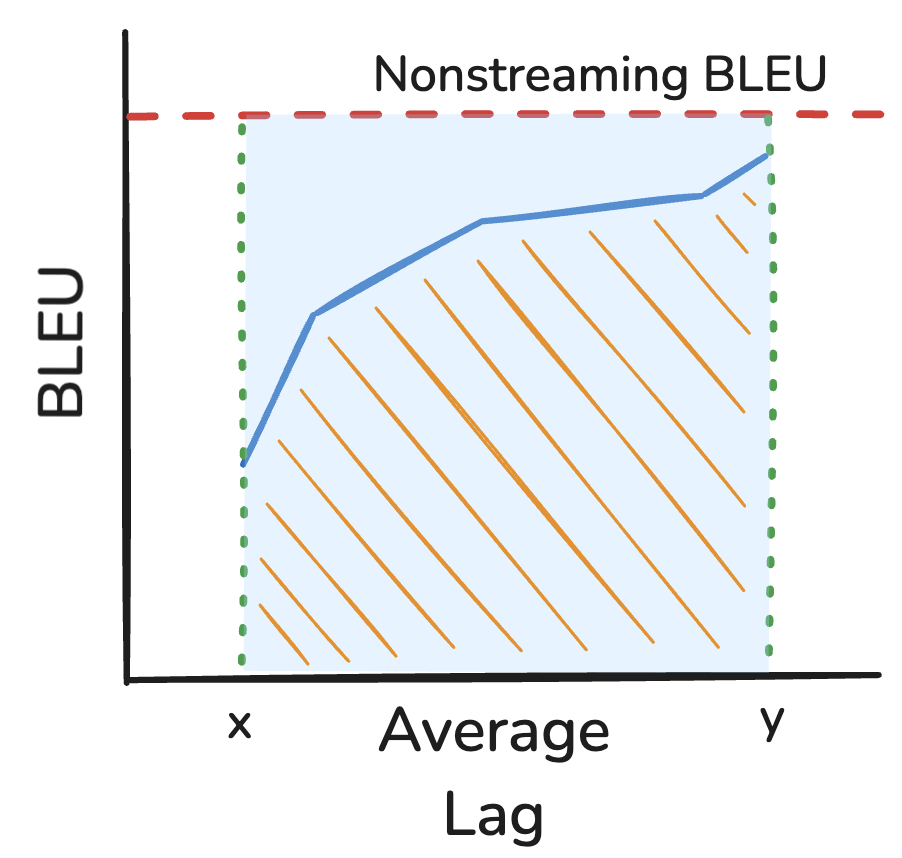}
    \caption{We define NoSE as the area of the orange shaded region divided by the area of the blue rectangle.}
    \label{fig:metric}
\end{figure}

To clarify our definition of the NoSE score, we provide a diagram in figure \ref{fig:metric}.

\section{Latency vs BLEU Evaluations}
\label{sec:al_bleu_graphs}

\begin{table*}[htbp] 
\fontsize{7}{8}\selectfont
\setlength{\tabcolsep}{4pt} 
\centering
\caption{Thresholds swept for generating AL vs. BLEU points for different models and datasets.}
\label{tab:thresholds_swept}
\begin{tabular}{@{}lll@{}} 
\toprule
Dataset & Model Name & Thresholds Swept \\
\midrule

\multirow{4}{*}{\textbf{MUST-C}} 
& Dig-SST (Original)        & Self-reported results \\
& DiSeg & Self-reported results \\
& EdAtt & Self-reported results \\
& Dig-SST (Our impl. MUST-C only)       & [.475, .5, .51, .52, .53, .54] \\
& REINA (MUST-C only) & [.935, .94, .9425, .945, .9475, .95] \\
& REINA (No Truncation) (MUST-C only) & [.975, .976, .977, .978, .979, .980] \\
& REINA                     & [.97, .975, .976, .977, .978, .979] \\
\midrule 

\multirow{3}{*}{\textbf{CVSS}} 
& REINA                     & [.97, .975, .976, .977, .978, .979, .98, .983, .985, .987] \\
& REINA w/o monotonicity    & [.976, .977, .978, .979, .98, .983, .985, .987] \\
& StreamSpeech              & Self-reported results \\ 
& SimulS2S-LLM              & Self-reported results \\ 

\bottomrule
\end{tabular}
\end{table*}

When evaluating our models on each dataset, we sweep across several policy thresholds. We find that the unique features of each dataset necessitate different thresholds to get the best results. In table \ref{tab:thresholds_swept} we show the list of thresholds we used for inferencing each model on each dataset. Unfortunately, the only we have to determine these thresholds is trial and error. We had to perform several offline inference sweeps across thresholds before finding a sweep that a) allowed us to compare to other works b) contained points in the AL/BLEU curve that were representative of how our model might be likely to be used in production.

\begin{figure}[htbp] 

    \begin{subfigure}[b]{0.45\textwidth} 
        \includegraphics[width=\linewidth]{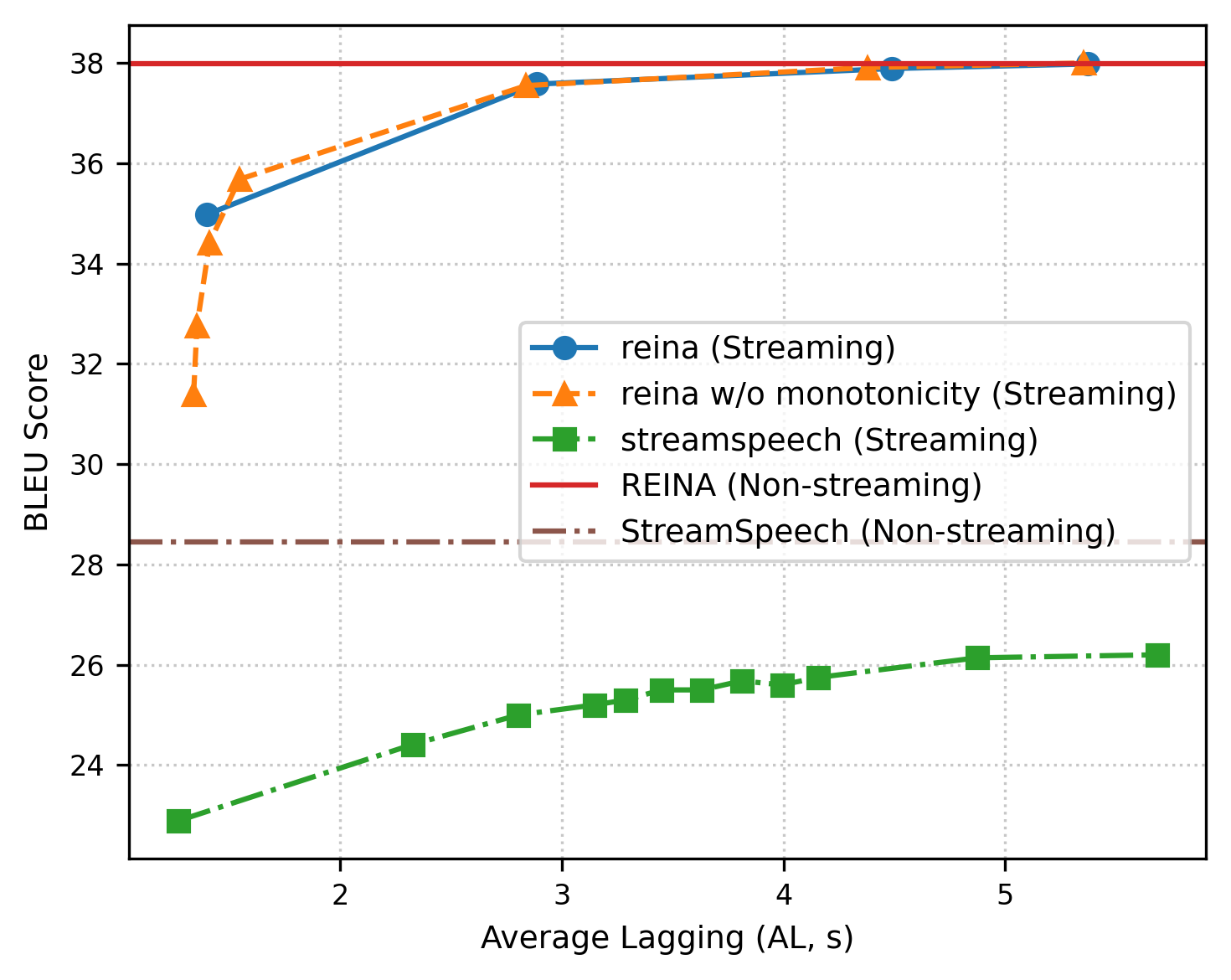}
        \caption{French $\rightarrow$ English (fr-en)} 
        \label{fig:cvss_fr_stacked} 
    \end{subfigure}
    

    \begin{subfigure}[b]{0.45\textwidth} 
        \includegraphics[width=\linewidth]{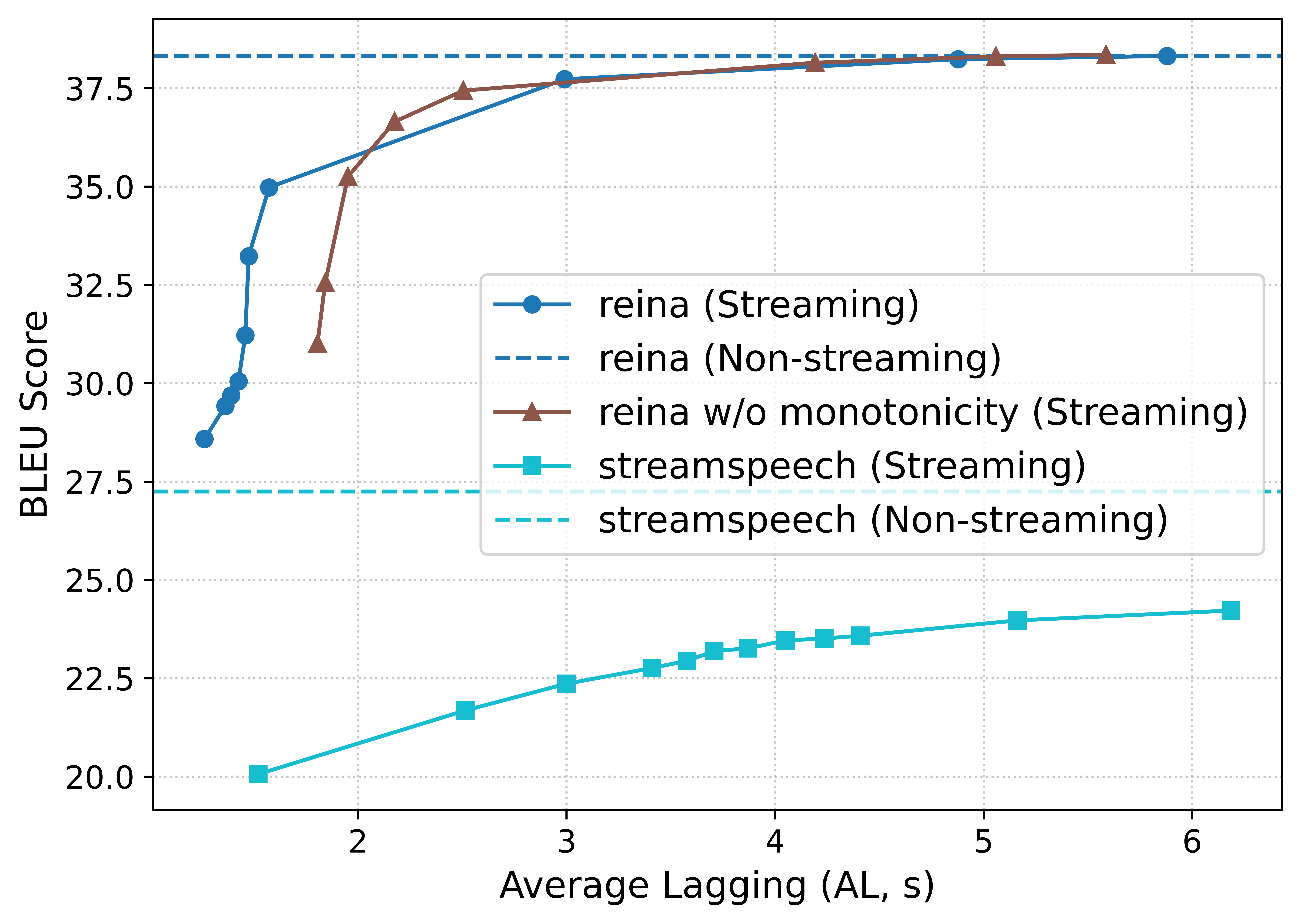}
        \caption{Spanish $\rightarrow$ English (es-en)} 
        \label{fig:cvss_es_stacked}
    \end{subfigure}

    \begin{subfigure}[b]{0.45\textwidth} 
        \includegraphics[width=\linewidth]{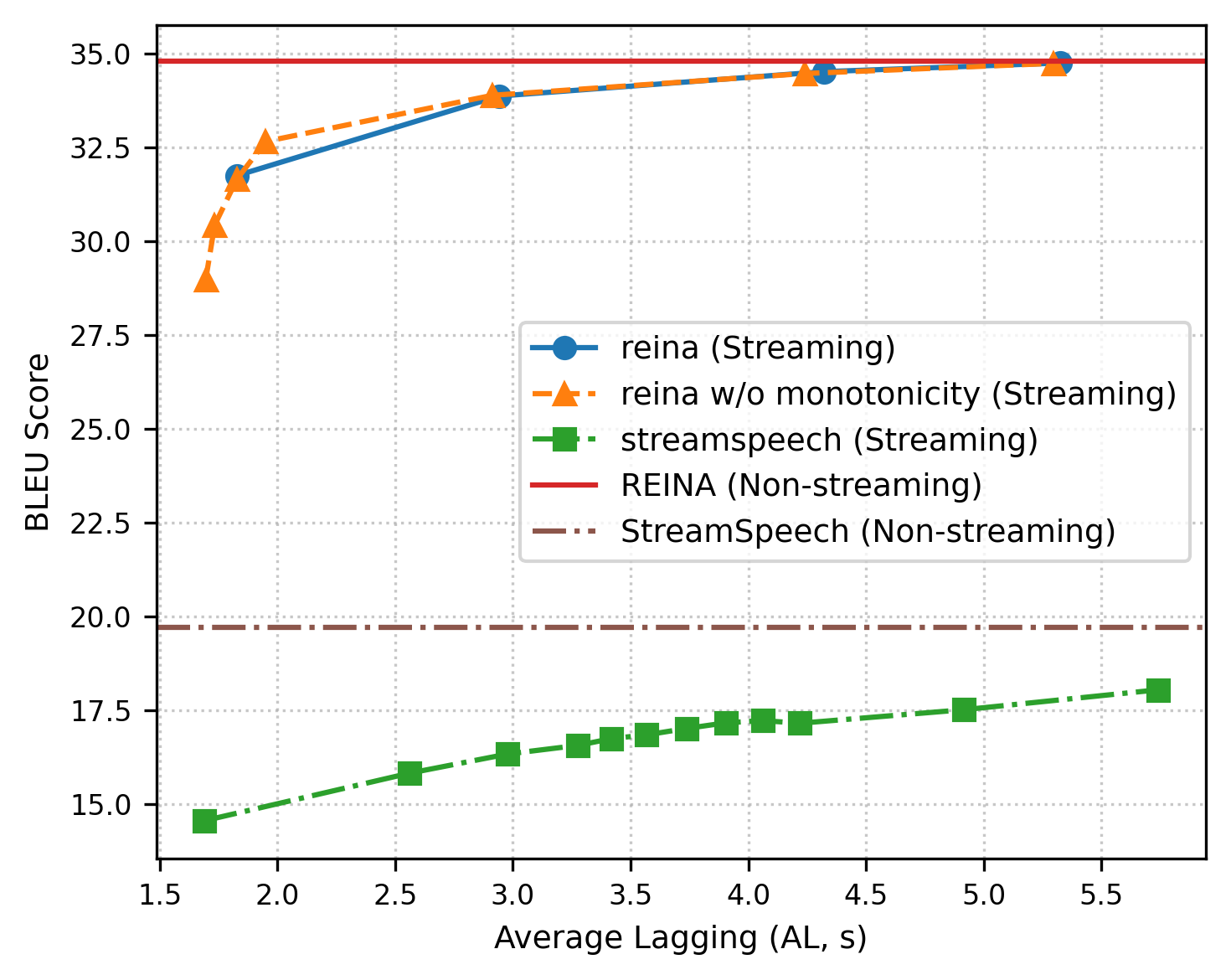}
        \caption{German $\rightarrow$ English (de-en)} 
        \label{fig:cvss_de_stacked}
    \end{subfigure}

    \caption{Average Lagging (AL) vs. BLEU score on CVSS-C. Dotted lines represent non-streaming BLEU scores. Note that StreamSpeech only reports ASR-BLEU in their paper, so we report StreamSpeech's ASR-BLEU rather than BLEU.} 
    \label{fig:cvss_graphs_all_stacked}
\end{figure}

\begin{figure}[htbp] 

    \begin{subfigure}[b]{0.45\textwidth} 
        \includegraphics[width=\linewidth]{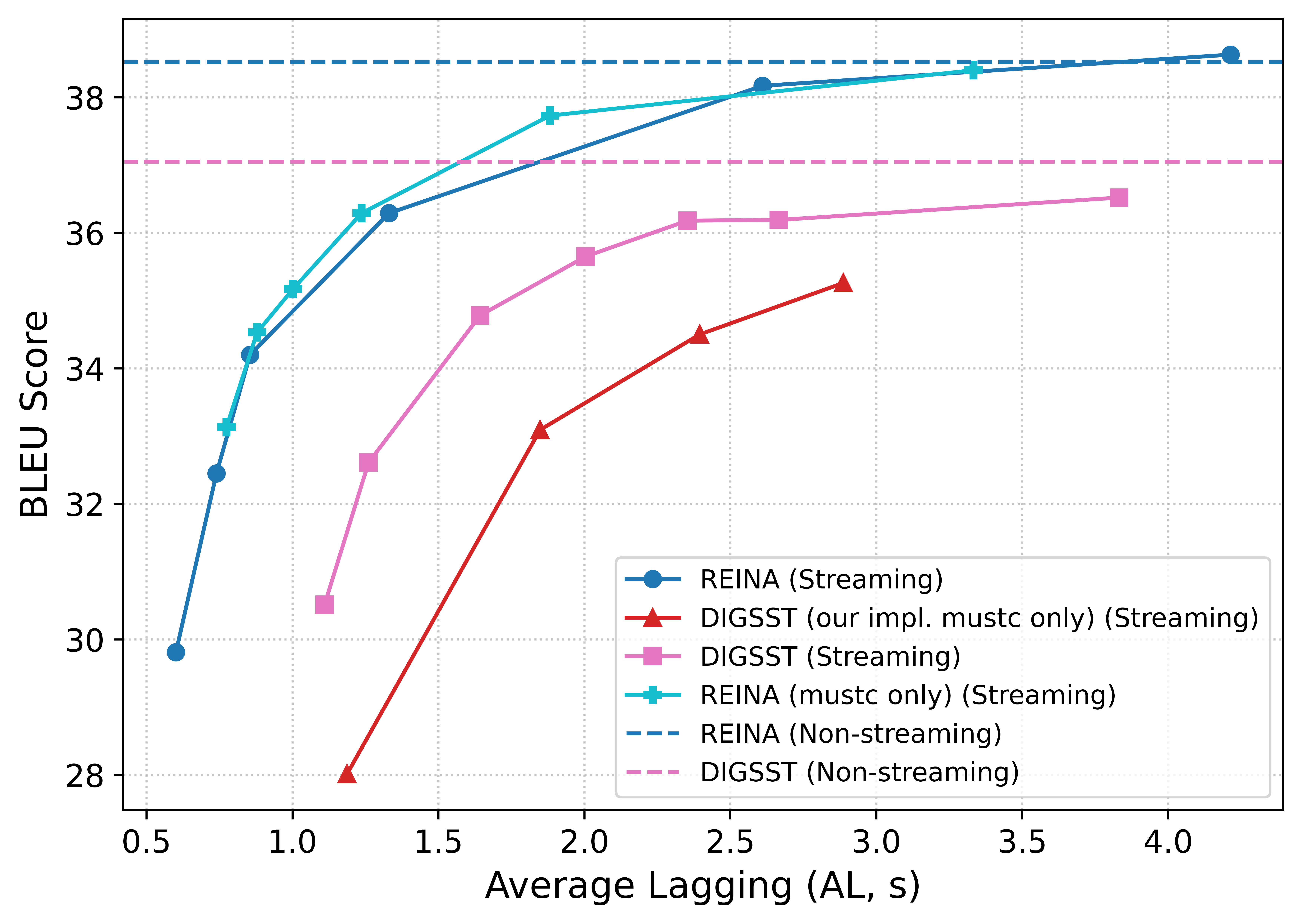}
        \caption{English $\rightarrow$ French (en-fr)} 
        \label{fig:mustc_fr_stacked} 
    \end{subfigure}
    

    \begin{subfigure}[b]{0.45\textwidth} 
        \includegraphics[width=\linewidth]{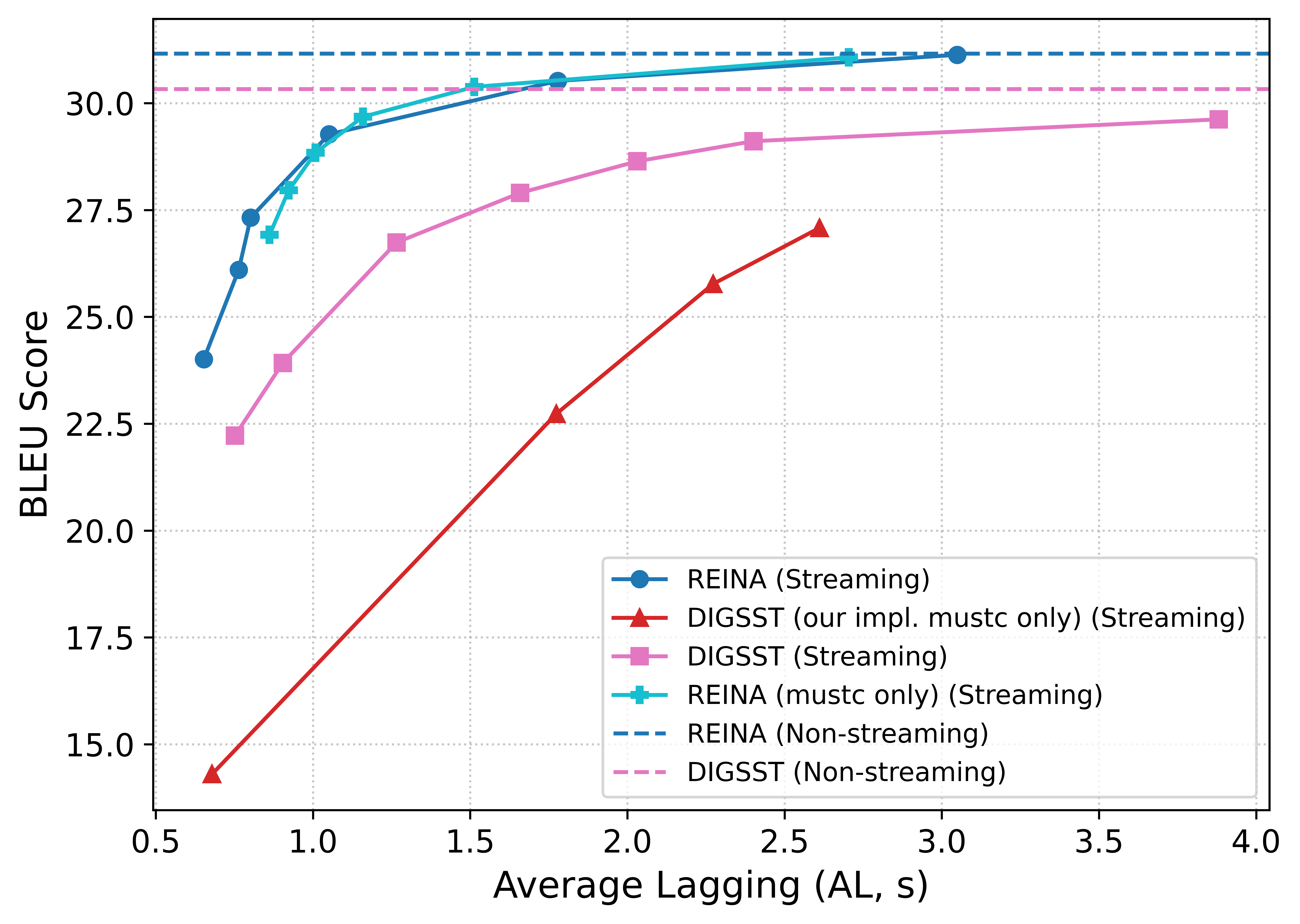}
        \caption{English $\rightarrow$ Spanish (en-es)} 
        \label{fig:mustc_es_stacked}
    \end{subfigure}

    \begin{subfigure}[b]{0.45\textwidth} 
        \includegraphics[width=\linewidth]{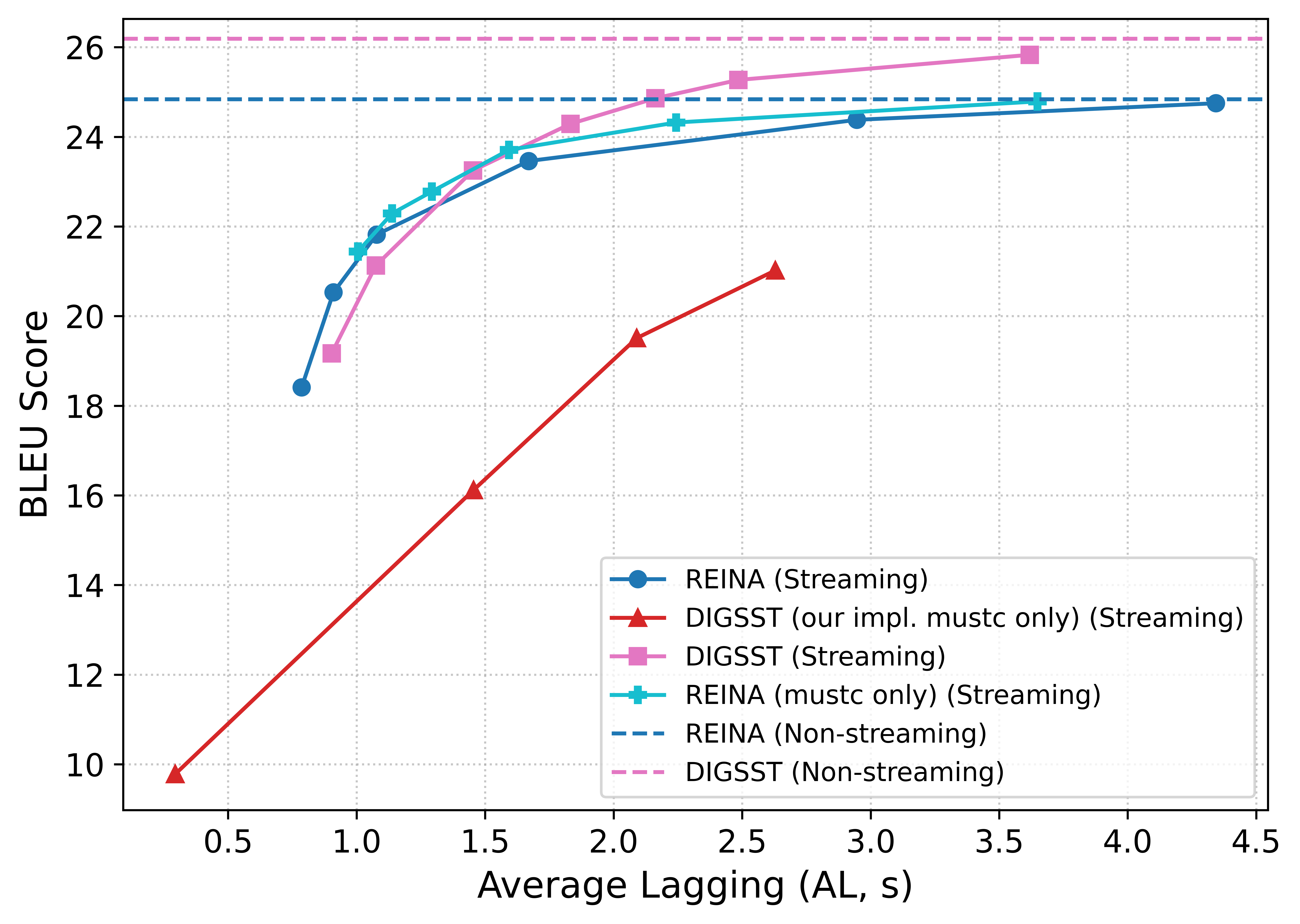}
        \caption{English $\rightarrow$ German (en-de)} 
        \label{fig:mustc_de_stacked}
    \end{subfigure}

    \caption{Average Lagging (AL) vs. BLEU score on MUST-C. Dotted lines represent non-streaming BLEU scores.} 
    \label{fig:mustc_graphs_all_stacked}
\end{figure}

We show AL vs BLEU graphs for all evaluations that appear in the paper in figures \ref{fig:cvss_graphs_all_stacked} and \ref{fig:mustc_graphs_all_stacked}.

\section{Hyperparameters}
\label{sec:hyperparameters}
We report model and dataset hyperparameters in table \ref{tab:hyperparameters}.

\begin{table}[htbp] 
\caption{Hyperparameters and Dataset Mixing Ratios used for training the S2TT model and the REINA policy network.}
\label{tab:hyperparameters}
\small 
\begin{tabular}{@{} >{\raggedright\arraybackslash}p{0.6\columnwidth} l @{}} 
\toprule
Parameter / Setting & Value \\
\midrule
\multicolumn{2}{@{}l}{\textbf{Text Decoder Configuration}} \\ 
\cmidrule(r){1-1} 
Dimension                                 & 512 \\
Attention Heads                           & 8 \\
Number of Layers                          & 16 \\ 
Feedforward Multiplier                    & 4 \\
Label Smoothing                         & 0.1 \\
Dropout                                 & 0.1 \\
\midrule
\multicolumn{2}{@{}l}{\textbf{Policy Network Configuration (REINA)}} \\ 
\cmidrule(r){1-1}
Monotonicity Epsilon ($\epsilon$)* & 0.5 \\ 
L2 Regularization Weight ($\lambda$)      & 0.05 \\ 
Number of Layers                          & 2 \\
Attention Heads                           & 4 \\
Feedforward Multiplier                    & 4 \\
Dimension                                 & 512 \\
\midrule

\multicolumn{2}{@{}l}{\textbf{Training Configuration}} \\ 
\cmidrule(r){1-1}
Learning Rate & 0.0001 (fixed) \\ 
Weight Decay      & 0.0001 \\ 
Gradient Clipping                          & 10.0 \\
Batch Size per Device & 16 \\
Gradient Accumulation Steps & 2 \\
Effective Batch Size & 768 \\
\midrule

\multicolumn{2}{@{}l}{\textbf{Dataset Mixing: Base S2TT Model Training (Stage 1 \& 2)}} \\ 
\cmidrule(r){1-1}
Dataset Name & Mixing Ratio \\ \cmidrule(r){1-2} 
MUST-C (train split)                      & 1 \\
CVSS (train split)                        & 1 \\
MLS (en $\rightarrow$ X)** & 2 \\ 
MLS (X $\rightarrow$ en)** & 4 \\ 
CCMatrix                                  & 4 \\
Mosel                                     & 4 \\
\midrule
\multicolumn{2}{@{}l}{\textbf{Dataset Mixing: REINA Policy Training (Stage 3)}} \\ 
\cmidrule(r){1-1}
Dataset Name & Mixing Ratio \\ \cmidrule(r){1-2} 
MUST-C (train split)                      & 2 \\
CVSS (train split)                        & 2 \\
MLS (en $\rightarrow$ X)** & 6 \\ 
MLS (X $\rightarrow$ en)** & 6 \\ 
\bottomrule
\multicolumn{2}{@{}p{\dimexpr\columnwidth-2\tabcolsep}@{}}{\footnotesize * Corresponds to $\epsilon$ in the monotonicity loss (Equation 3).} \\ 
\multicolumn{2}{@{}p{\dimexpr\columnwidth-2\tabcolsep}@{}}{\footnotesize ** Derived from Multilingual Librispeech (MLS) dataset (see Section 3.5).} \\ 
\end{tabular}
\end{table} 

\section{EMMA Discussion}
\label{sec:emma}
In our paper, we touch on the complexities of implementing streaming via monotonic attention as in the EMMA method used in \cite{communication2023seamlessmultilingualexpressivestreaming}. In this section, we wish to provide some more background on the computational and numerical challenges that arise from training with EMMA.

In our research, we implemented EMMA from scratch on top of a non-streaming REINAStream model. The train-time computations for EMMA requires computing a matrix of size [batch\_size $\times$ attention\_heads $\times$ num\_text\_tokens $\times$ audio\_sequence\_length $\times$ audio\_sequence\_length] within each cross-attention layer of the decoder. Seeing as the audio sequence length coming out of the whisper encoder is 1500, we use 8 attention heads, and a short token sequence may contain around 25 tokens, using fp32 precision, we require 2GB of VRAM for a single cross-attention layer at batch size 1. Ultimately, we were only able to train EMMA with batch size 1 and truncating the encoder output sequence length to 500, despite using A100-80G GPUs. This made for very slow, expensive training.

Furthermore, the EMMA estimation requires computing a cumulative product across the audio sequence length dimension. A cumulative product of 500 small floating point values is numerically unstable and often results in rounding to 0. Perhaps the original authors used a sum of log values instead.

Lastly, as EMMA computes a separate policy for every attention head of every cross-attention layer, it is unclear which one to use for the final inference policy. In the public inference code associated with \cite{communication2023seamlessmultilingualexpressivestreaming}, the layer to use for the policy is simply taken as an argument. Attention heads from that layer are aggregated via a max, min, or mean. Empirically, we found that some layers and heads learned useful policies, while the majority did not.

Due to these challenges implementing and training EMMA, we did not include it in our evaluations and pursued simpler streaming methods, motivating us to invent REINA.

We also note that many of the same issues arise with training neural transducer~\cite{graves2012sequencetransductionrecurrentneural} models. We also spent considerable effort implementing transducers for SimulST. As with EMMA, we found transducer methods to be excessively expensive to train and very hard to make converge.

\section{FLEURS Evaluation}
\label{sec:fleurs}
In this section, we present an attempt at comparing to Seamless on FLEURS. We construct the FLEURS dataset for en $\rightarrow$ {de, en, fr} and {de, en, fr} $\rightarrow$ en, deduplicating on unique source audio and target language. We trim the dataset using the same Silero VAD Seamless uses. When inferencing, we soon noticed that as our model is trained entirely on audios not trimmed with VAD, it tends to expect an ending silence and over-generates, resulting in decreased BLEU scores. We also attempted inferencing Seamless on our copy of FLEURS using their open source code, but ran into implementation difficulties due to scarce documentation.

In a final attempt at a comparison, we use Seamless' reported scores from their paper. We use VAD to trim our audios and then augment them with 2 seconds of white noise at the end to help our model. We tested this trick without VAD on our other datasets and found it has a negative impact on BLEU, but we found it was still an improvement over inferencing on VAD-trimmed audios on FLEURS.

\begin{figure}[htbp] 

    \begin{subfigure}[b]{0.45\textwidth} 
        \includegraphics[width=\linewidth]{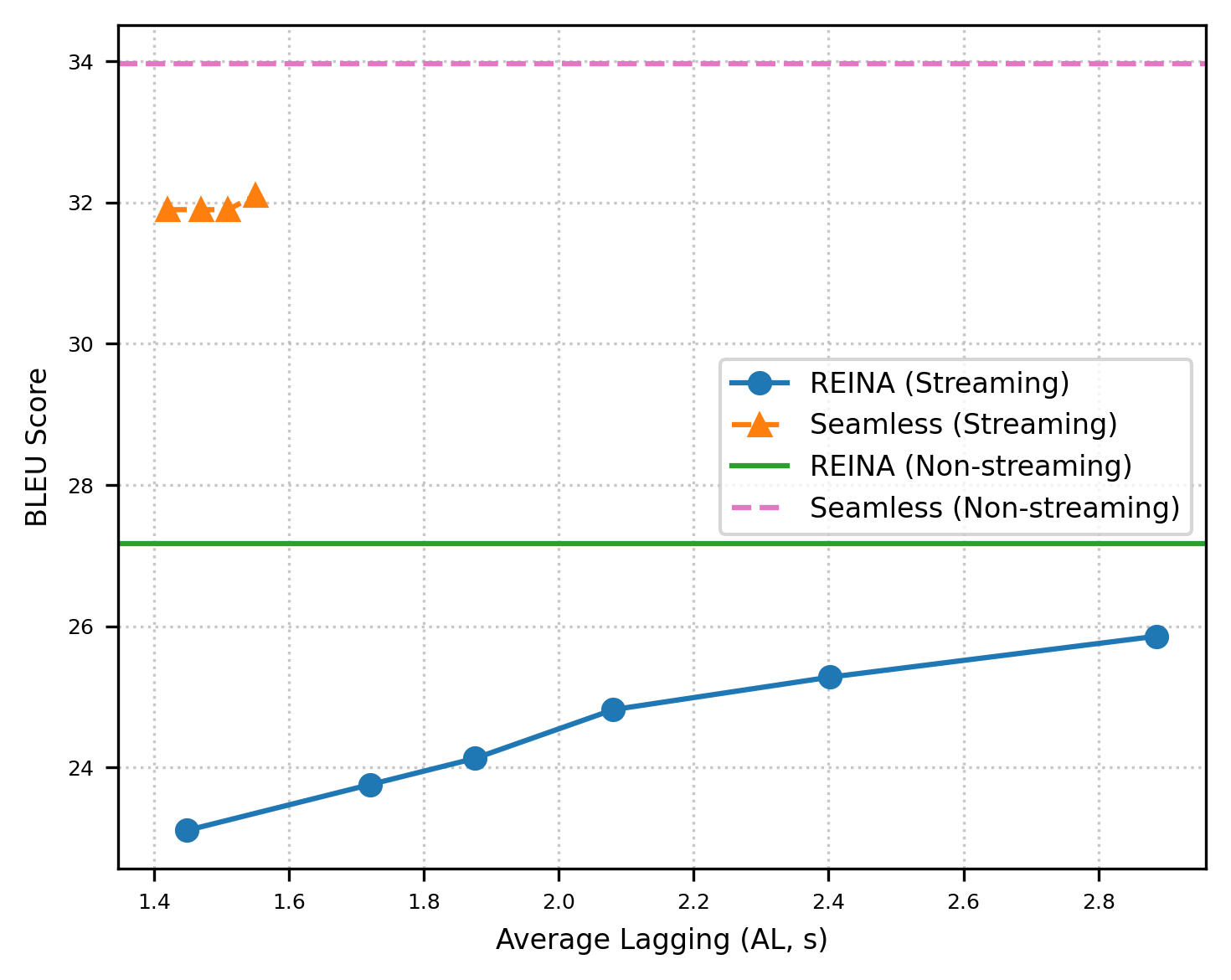}
        \caption{French $\rightarrow$ English (fr-en)} 
    \end{subfigure}
    

    \begin{subfigure}[b]{0.45\textwidth} 
        \includegraphics[width=\linewidth]{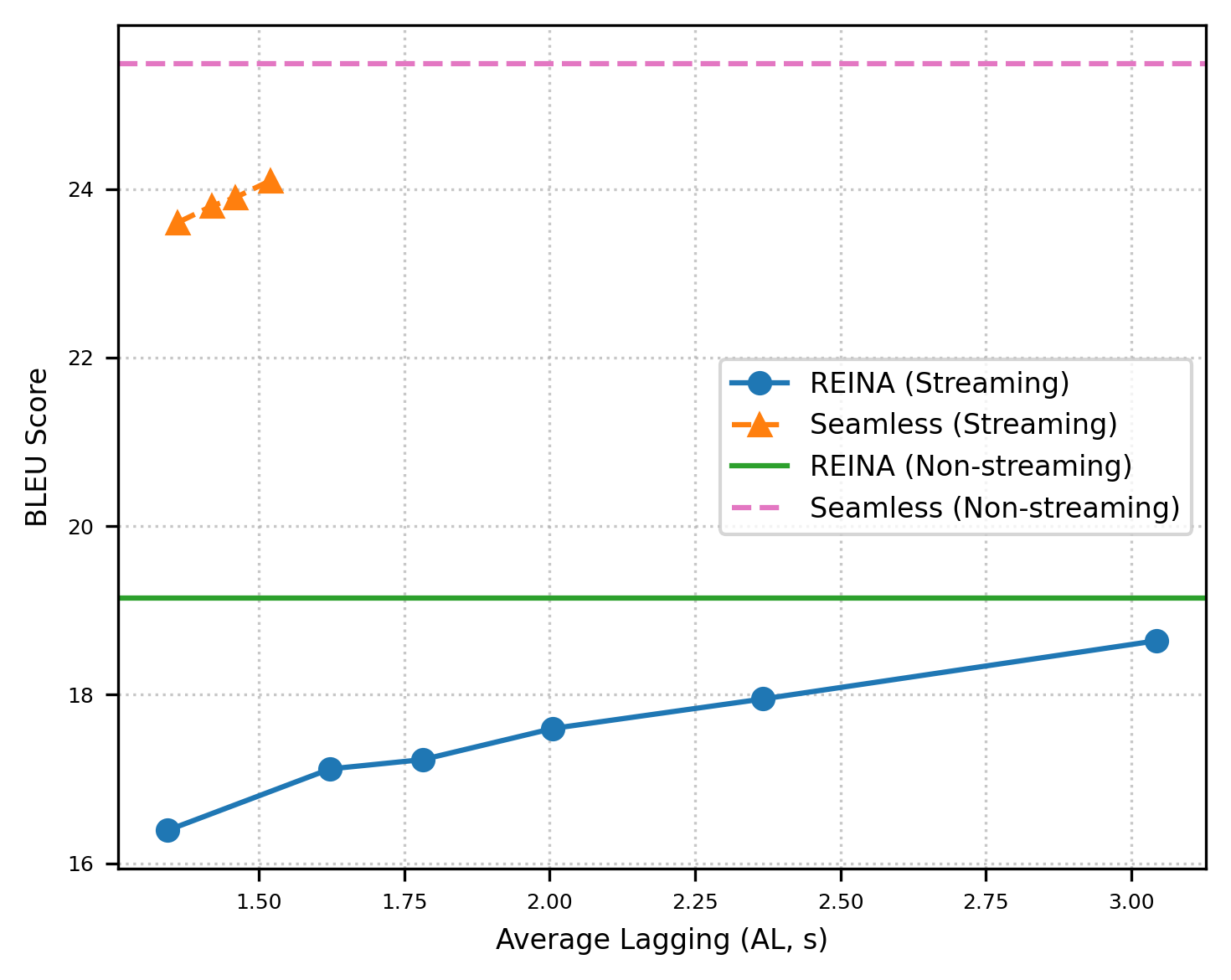}
        \caption{Spanish $\rightarrow$ English (es-en)} 
    \end{subfigure}

    \begin{subfigure}[b]{0.45\textwidth} 
        \includegraphics[width=\linewidth]{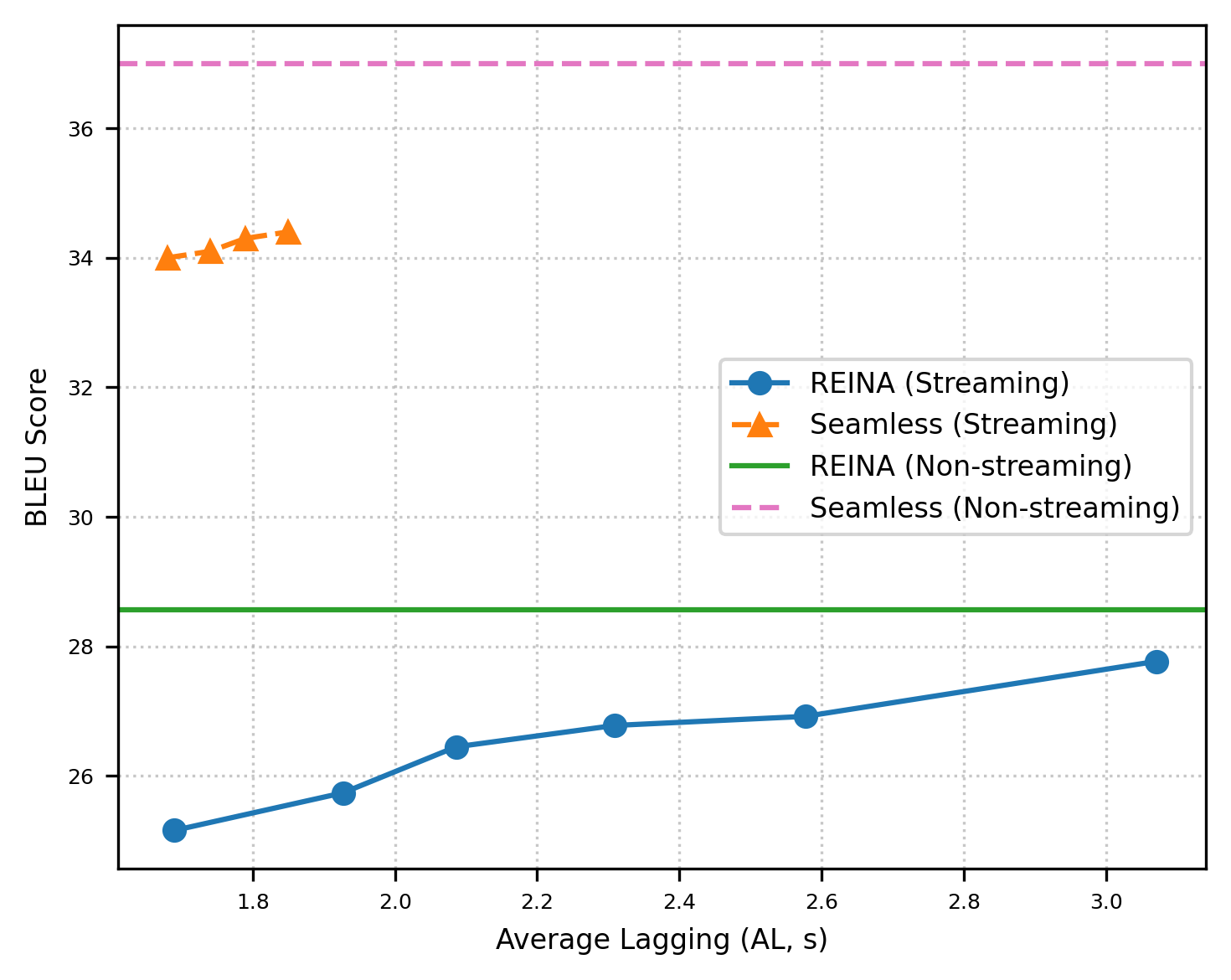}
        \caption{German $\rightarrow$ English (de-en)} 
    \end{subfigure}

    \caption{Average Lagging (AL) vs. BLEU score on FLEURS X $\rightarrow$ en. Dotted lines represent non-streaming BLEU scores.} 
    \label{fig:fleurs_xtoen_stacked}
\end{figure}

\begin{figure}[htbp] 

    \begin{subfigure}[b]{0.45\textwidth} 
        \includegraphics[width=\linewidth]{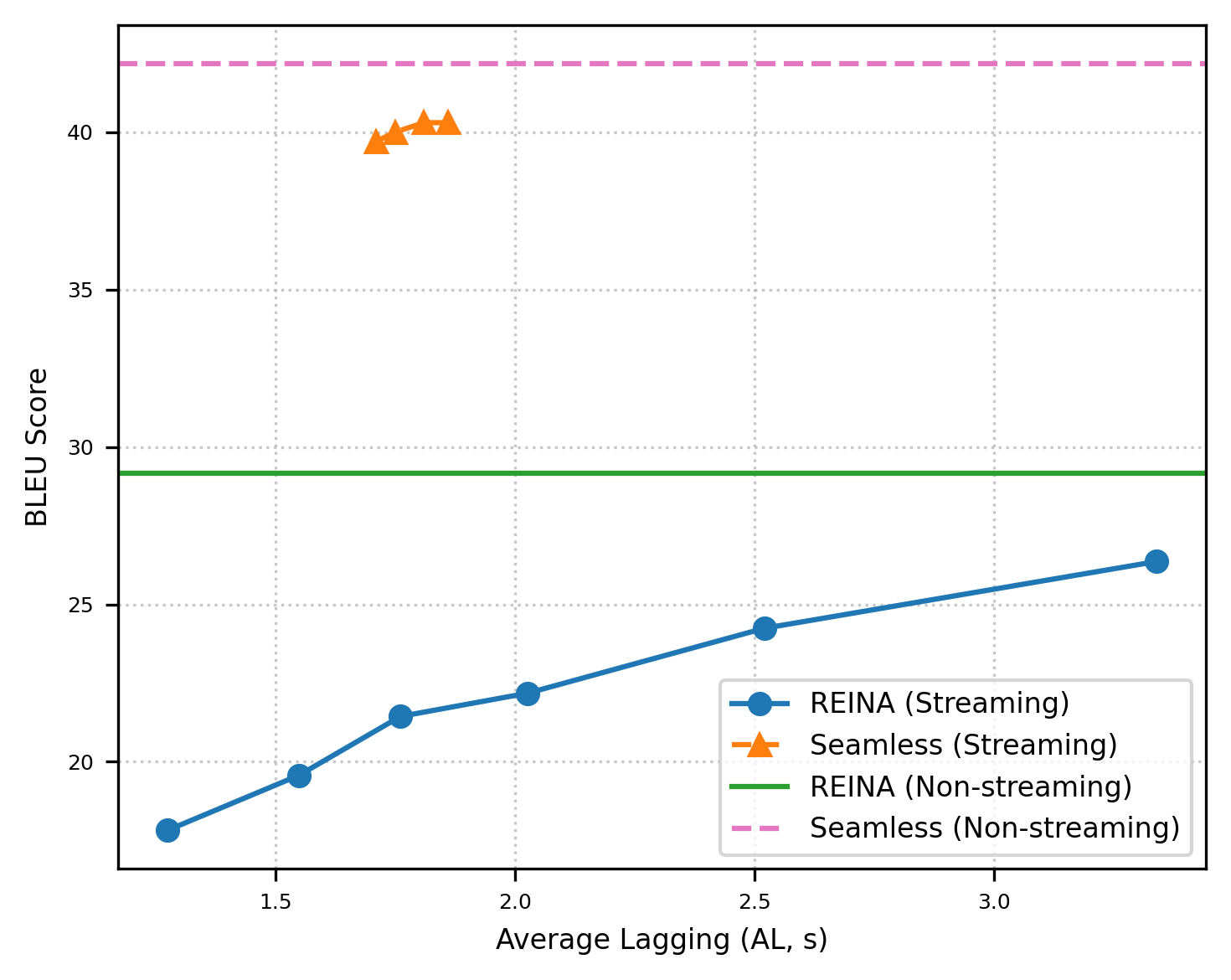}
        \caption{English $\rightarrow$ French (en-fr)} 
    \end{subfigure}
    

    \begin{subfigure}[b]{0.45\textwidth} 
        \includegraphics[width=\linewidth]{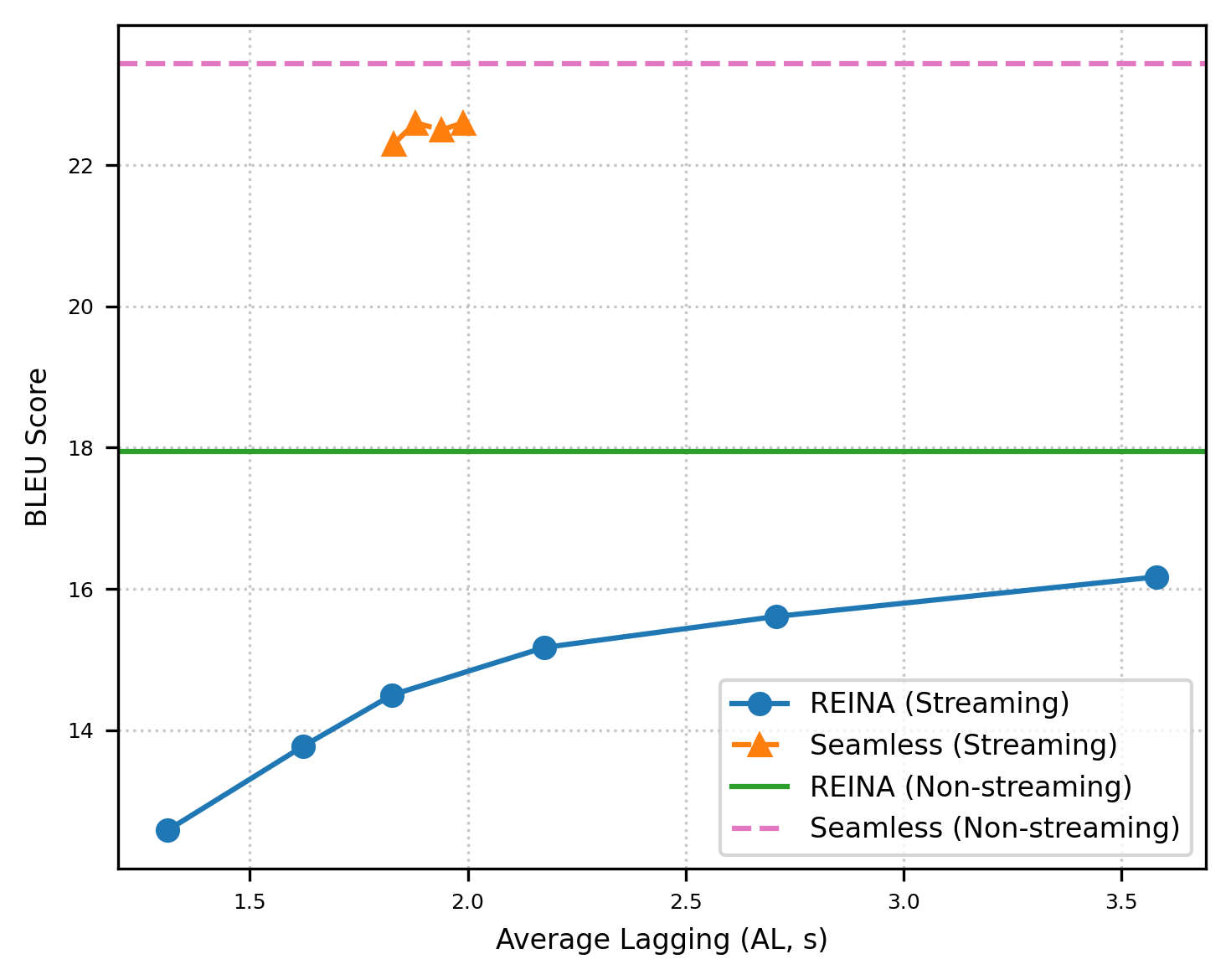}
        \caption{English $\rightarrow$ Spanish (en-es)} 
    \end{subfigure}

    \begin{subfigure}[b]{0.45\textwidth} 
        \includegraphics[width=\linewidth]{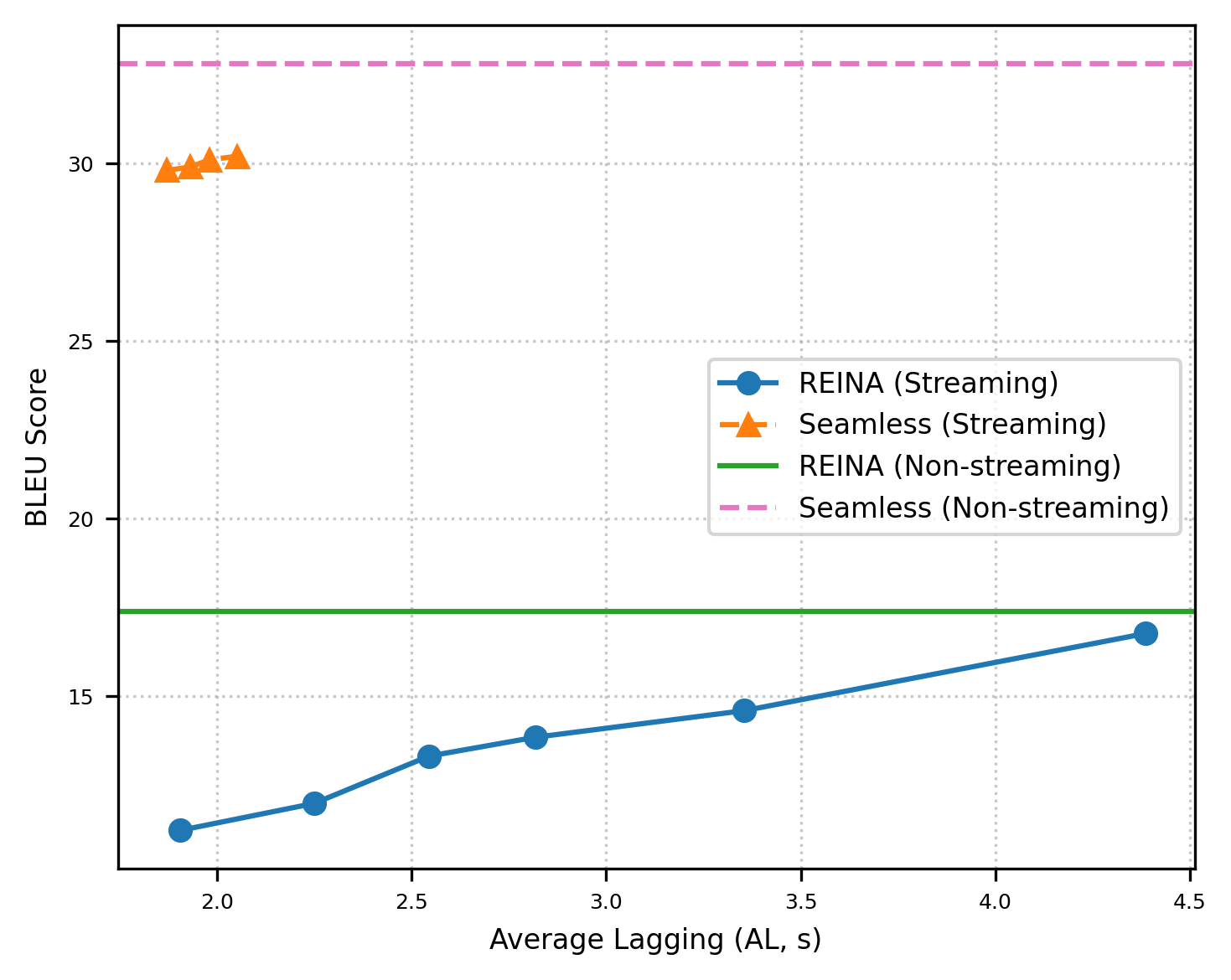}
        \caption{English $\rightarrow$ German (en-de)} 
    \end{subfigure}

    \caption{Average Lagging (AL) vs. BLEU score on FLEURS en $\rightarrow$ X. Dotted lines represent non-streaming BLEU scores.} 
    \label{fig:fleurs_entox_stacked}
\end{figure}

We show NoSE scores in table \ref{tab:fleurs_results}, graphs for the X $\rightarrow$ en direction in figure \ref{fig:fleurs_xtoen_stacked}, and graphs for the en $\rightarrow$ X direction in figure \ref{fig:fleurs_entox_stacked}. As our model is much smaller than Seamless, our non-streaming BLEU is unsurprisingly much lower. Of more interest are the NoSE scores, which show we are comparable to Seamless on the X $\rightarrow$ en directions but worse on en $\rightarrow$ X. We believe this difference is primarily accounted for by the aforementioned VAD issues. We also note that REINA is vastly cheaper and easier to train than Seamless' EMMA streaming method (see appendix \ref{sec:emma}), meaning that even with comparable streaming quality, REINA is in most cases preferable to EMMA.

\begin{table*}[h] 

\fontsize{6}{7}\selectfont
\setlength{\tabcolsep}{3pt} 
\centering

\begin{tabular}{@{}llcccccc@{}} 
\toprule
Dataset & Item / Model & en$\rightarrow$de & en$\rightarrow$fr & en$\rightarrow$es & de$\rightarrow$en & fr$\rightarrow$en & es$\rightarrow$en \\
\midrule

\multirow{5}{*}{\textbf{FLEURS}} & \itshape Bounds $[x,y]$ & \scriptsize [1.87, 2.05] & \scriptsize [1.71, 1.86] & \scriptsize [1.83, 1.99] & \scriptsize [1.68, 1.85] & \scriptsize [1.42, 1.55] & \scriptsize [1.36, 1.52] \\
\cmidrule(lr){2-8} 
& Seamless        & \textbf{0.914} & \textbf{0.951} & \textbf{0.960} & .924 & \textbf{.940} & \textbf{.936} \\
& REINAStream       & 0.896 & 0.861 & 0.916 & \textbf{.943} & .866 & \textbf{.936} \\


\bottomrule
\end{tabular}%
\caption{Comparison of NoSE $(\uparrow)$ values on FLEURS for Seamless and REINAStream.}
\label{tab:fleurs_results}

\end{table*}

\end{document}